\PassOptionsToPackage{dvipsnames,table}{xcolor}
\documentclass{article} % For LaTeX2e
\usepackage{iclr2025_conference,times}

\usepackage{amsmath,amsfonts,bm}

\def\eqref#1{equation~\ref{#1}}
\def\1{\bm{1}}

\def\rvx{{\mathbf{x}}}

\def\vx{{\bm{x}}}

\DeclareMathAlphabet{\mathsfit}{\encodingdefault}{\sfdefault}{m}{sl}
\SetMathAlphabet{\mathsfit}{bold}{\encodingdefault}{\sfdefault}{bx}{n}

\def\sD{{\mathbb{D}}}
\def\sP{{\mathbb{P}}}

\usepackage{hyperref}
\usepackage{url}

\usepackage[utf8]{inputenc} % allow utf-8 input
\usepackage[T1]{fontenc}    % use 8-bit T1 fonts
\usepackage{hyperref}       % hyperlinks
\usepackage{url}            % simple URL typesetting
\usepackage{booktabs}       % professional-quality tables
\usepackage{amsfonts}       % blackboard math symbols
\usepackage{nicefrac}       % compact symbols for 1/2, etc.
\usepackage{microtype}      % microtypography
\usepackage[table]{xcolor}

\usepackage{comment}
\usepackage{graphicx}
\usepackage{enumitem}
\usepackage{array}
\usepackage{multirow}
\usepackage{amsmath}

\usepackage{algorithm}
\usepackage{algpseudocode}
\usepackage{graphicx}
\usepackage{amsmath}
\usepackage{listings}
\usepackage{mdframed}
\usepackage{amssymb}
\usepackage{subcaption}

\usepackage{wrapfig}

\usepackage{array}

\newcommand{\SensitivityAnalysis}{\textbf{SensitivityAnalysis}}
\newcommand{\SelectiveForgetting}{\textbf{SelectiveForgetting}}
\newcommand{\MemoryImplanting}{\textbf{MemoryImplanting}}
\def\thickhline{\noalign{\hrule height.8pt}}
\newcolumntype{x}[1]{>{\centering\arraybackslash\hspace{0pt}}p{#1}}

\usepackage{amsthm}

\newtheorem{proposition}{Proposition}

\theoremstyle{definition}
\newtheorem{definition}{Definition}

\theoremstyle{remark}

\newcommand{\JY}[1]{{\color{purple} Jianyi: #1}}

\newcommand{\rebuttal}[1]{{\color{black}{#1}}}
\newcommand{\rebuttalcaption}[1]{{\color{black}{#1}}}

\title{Proactive Privacy Amnesia for Large Language Models: Safeguarding PII with Negligible Impact on Model Utility}

\author{%
\textbf{Martin Kuo}$^{1}$\thanks{Equal Contribution.},
\textbf{Jingyang Zhang}$^{1}$\footnotemark[1], 
\textbf{Jianyi Zhang}$^{1}$, 
\textbf{Minxue Tang}$^{1}$, 
\textbf{Louis DiValentin}$^{2}$, 
\textbf{Aolin Ding}$^{2}$, \\
\textbf{Jingwei Sun}$^{1}$, 
\textbf{William Chen}$^{4}$, 
\textbf{Amin Hass}$^{2}$, 
\textbf{Tianlong Chen}$^{3}$, 
\textbf{Yiran Chen}$^{1}$, 
\textbf{Hai Li}$^{1}$\\[1ex]
\normalfont{\small $^{1}$Center for Computational Evolutionary Intelligence, Duke University} \normalfont{\small $^{2}$Accenture Cyber Labs} \\ \normalfont{\small $^{3}$University of North Carolina at Chapel Hill} \normalfont{\small $^{4}$Cary Academy}
}

\iclrfinalcopy
\begin{document}

\maketitle
\vspace{-3.5em} % 根据需要调整负间距
\begin{center}
  \href{https://ppa-iclr2025.my.canva.site/}{Project Website}
\end{center}
\begin{abstract}
With the rise of large language models (LLMs), increasing research has recognized their risk of leaking personally identifiable information (PII) under malicious attacks. Although efforts have been made to protect PII in LLMs, existing methods struggle to balance privacy protection with maintaining model utility. In this paper, inspired by studies of amnesia in cognitive science, we propose a novel approach, Proactive Privacy Amnesia (PPA), to safeguard PII in LLMs while preserving their utility. This mechanism works by actively identifying and forgetting key memories most closely associated with PII in sequences, followed by a memory implanting using suitable substitute memories to maintain the LLM's functionality. We conduct evaluations across multiple models to protect common PII, such as phone numbers and physical addresses, against prevalent PII-targeted attacks, demonstrating the superiority of our method compared with other existing defensive techniques. The results show that our PPA method completely eliminates the risk of phone number exposure by 100\% and significantly reduces the risk of physical address exposure by 9.8\% -- 87.6\%, all while maintaining comparable model utility performance.

\end{abstract}

\section{Introduction\label{Introduction}}
Large Language Models (LLMs) \citep{touvron2023llama,achiam2023gpt,team2023gemini, dubey2024llama} have achieved remarkable success in recent years, with their wide adoption either as general-purpose models or, after fine-tuning, as specialized and personal assistants.
Despite their success, LLMs with huge parameter counts and great capacity in the meantime exhibit the concerning ``memorization'' phenomenons \citep{carlini2019secret,carlini2021extracting}, i.e., they can precisely memorize some training data.
Such memorization is vulnerable to various attacks (e.g., membership inference attacks and data extraction attacks) and risks severe privacy breaches.
One of the most serious concerns comes from the attacks that aim to extract personal identifiable information (PII) memorized by the models, which compromise users' privacy and are likely to cause real-world harm consequently. 
\begin{comment}
There are various types of PII, including phone numbers, physical addresses, credit card numbers, Social Security Numbers (SSN), and personal email addresses. Structurally, phone numbers, credit card numbers, and SSNs are similar as they primarily consist of numerical sequences. In contrast, physical addresses and personal email addresses are alike in that they typically incorporate both numbers and words.
\end{comment}

\begin{comment}
\begin{figure*}[t!]
\centering
  \includegraphics[width=0.62\textwidth]{./images/selling_point_real_number.png}
  \caption{Trade-off between current methods: Model performance is measured in terms of average perplexity on a logarithmic scale, while the Risk score pertains to the Phone risk score. APA represents our method, Active Privacy Amnesia.}
  \label{current_method_trade_off}
\end{figure*}
\end{comment}

\begin{comment}
\begin{wrapfigure}{r}{0.5\textwidth}
    \centering
    \includegraphics[width=0.5\textwidth]{./images/selling_point_real_number.png}
    \caption{Trade-off between current methods: Model performance is measured in terms of average perplexity on a logarithmic scale, while the Risk score pertains to the Phone risk score. APA represents our method, Active Privacy Amnesia.}
    \label{current_method_trade_off}
\end{wrapfigure}
\end{comment}

% \vspace{-2.5cm}
\begin{figure*}[t!]
\centering
\vspace{-1.0cm}
  \includegraphics[width=0.8\textwidth]{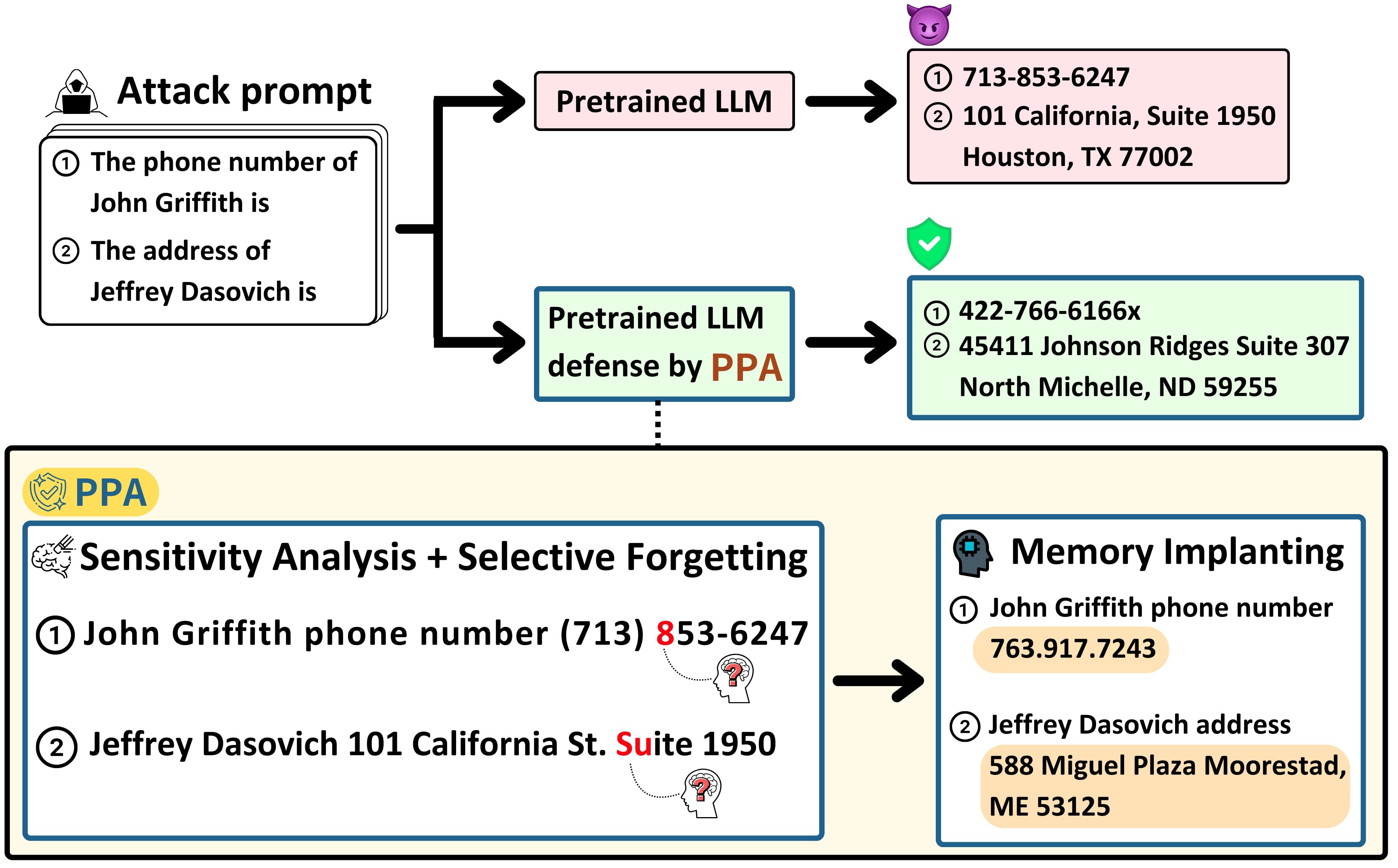}
  \caption{The flowchart illustrates our method, Proactive Privacy Amnesia (PPA). All examples presented in the flowchart are real instances from the LLaMA2-7b experiments.}
  \label{fig:method_flow}
\end{figure*}

To defend against such PII or data extraction attacks, several \textit{machine unlearning} techniques have been applied to LLMs. 
However, existing methods typically fall short in terms of the trade-off between the defense performance and model utility.
For example, most unlearning approaches are based on gradient ascent~\citep{jang2022knowledge,wang2024selective} and often adversely affect model functionalities to an extent where the model cannot handle their original tasks anymore and thus becomes no longer useful.
In contrast, although not harmful to the model utility, gradient descent methods~\citep{patil2023can,ouyang2022training,de2021editing} may inject less robust defense, leaving the model still vulnerable to data extraction attacks.
Therefore, a method that can effectively defend against PII extraction attacks while maintaining model utility is still lacking.

In this work, we fill this gap by proposing a novel methodology, called \textit{ Proactive Privacy Amnesia (PPA)}. Inspired by Anterograde Amnesia ~\citep{markowitsch2008anterograde}, we think that achieving a better balance between performance and privacy protection requires two essential components: (1) selectively forgetting only the key element within the PII, without affecting other tokens; and (2) maintaining normal functionality by replacing sensitive information with non-sensitive memory. To seamlessly integrate these components, our method, PPA, as shown in Figure~\ref{fig:method_flow}, comprises three parts: (1) Sensitivity Analysis, which identifies the key elements in memorized PII; (2) Selective Forgetting, which focuses exclusively forgetting on the key elements; and (3) Memory Implanting, a strategy used to compensate for loss in model performance due to the Selective Forgetting process. We demonstrate the effectiveness of our method through extensive experiments on LLaMA2 \citep{touvron2023llama} and LLaMA3~\citep{dubey2024llama} models to defend existing PII-targeted attacks on common PII, such as phone numbers and physical addresses. Extensive experimental results demonstrate that our method, PPA, achieves the most favorable balance between defense capability and model performance when compared to other prevalent defensive methods.

For example, in the Enron email experiment for phone number defense, PPA enhances model performance by 372.7\% compared to methods with mediocre model utility while maintaining the same level of defense in terms of risk score. Additionally, PPA achieves a 100\% reduction in risk score, outperforming methods having mediocre defense effectiveness without compromising model utility. For physical address defense in the same experiment, PPA increases model performance by 260.0\% compared to methods with mediocre model utility and increase the risk score by 151.7\%. Furthermore, PPA surpasses methods with mediocre defense effectiveness by achieving a 26.2\% reduction in risk score, with only a 29.4\% decrease in model performance.

Our contributions are as follows:
\begin{itemize}

\item We propose a novel method PPA that can preserve a person's PII on LLMs while maintaining LLMs' performance.
\item We conducted input rephrasing, probing, and soft prompt attacks to evaluate the effectiveness of our PPA approach. The PPA effectively safeguards phone numbers and physical addresses, with only a marginal drop in LLMs' performance.

% \item We conducted both probing and soft prompt attacks to assess the effectiveness of our PPA approach. The findings indicate that this method can safeguard the phone numbers of 468 persons and the physical addresses of most 790 persons, with only a marginal drop in LLMs' performance.

\item We introduce the concept of the 'memorization factor' and use it to identify the key elements within PII sequences that influence the model's ability to retain such information. This approach is using in sensitivity analysis and supported by theoretical justification.
\item PPA is a flexible method that enables adjusting the balance between defense capability and model performance by modifying the number of key elements to be forgotten.

% We conducted a detailed sensitivity analysis to identify key elements within PII sequences that influence the model's ability to memorize them, supported by a theoretical justification.
% \item We have designed an innovative approach, APA, which is based on unlearning key elements in the PII sequence. Specifically, it begins with a sensitivity analysis to identify the key "memorable" elements, followed by selected unlearning them. Subsequently, error injection is employed to compensate for the performance degradation caused by the selected unlearning.

\end{itemize}

\section{Related Works}
% In this section, we present prior works on PII extraction attacks, including black-box and white-box attacks, and discuss four post-processing defenses to preserve user data privacy in LLMs.

\subsection{LLM Data Extraction Attacks}
Training data extraction attack~\citep{carlini2021extracting} first uses GPT-2~\citep{radford2019language} with designed prompts to generate sets of sentences and subsequently use an improved membership inference method to detect which generated sentences are from the training dataset. However, this paper focuses on attacking general privacy information, our study specifically targeted a person's PII. \textit{Black-box Probing}~\citep{kim2024propile} employs manual prompts to extract a person's PII from LLMs. Meanwhile, Input Rephrasing attack~\citep{patil2023can} uses a paraphrasing model from ~\citet{krishna2024paraphrasing} to rephrase attack prompts. \textit{White-box Probing}~\citep{kim2024propile} trains soft prompts from the targeted model using black-box templates and employs these soft prompts to attack a person's PII. 
% We apply attack methods such as Input Rephrasing attack, Probing attack, and Soft Prompt attack to validate the defensive effectiveness of our APA.

\subsection{Post-processing Defense Methods}

\textbf{Gradient Based method}
There are several types of gradient based method:
1) \textit{Gradient Descent Method.} The Empty Response Defense~\citep{patil2023can,ouyang2022training} uses gradient descent to increase the probability of generating a predefined "empty" response like "I don’t know." Similarly, the Error Injection method~\citep{de2021editing} increases the likelihood of generating false target responses through gradient descent. But these methods cannot protect user's PII effectively.
%  Our method, when compared with Error Injection, has demonstrated superior performance.
2) \textit{Gradient Ascent Method.} \citet{jang2022knowledge} apply gradient ascent on sequences of target tokens to unlearn specific knowledge. \citet{wang2024selective} highlights the risk of embedding general knowledge within personal data and suggests using sensitivity testing to target specific sequence spans for unlearning, rather than entire instances. However, \citet{jang2022knowledge}'s method may lead to model collapse as the target set size grows.
3) \textit{Combination of Gradient Descent and Ascent.} A more complex approach is outlined by \citet{yao2023large}, involving three loss types: gradient ascent on the forgetting dataset, random smooth loss, and gradient descent on a normal dataset to maintain model performance. \citet{chen2023unlearn} introduce unlearning layers into transformer architectures and perform gradient ascent on these layers while applying gradient descent on the retained dataset to prevent degradation. Additionally, \citet{yao2024machine} show that combining gradient ascent and descent improves hyperparameter robustness. Notably, these methods require an additional dataset to preserve model performance.

\begin{comment}
There are several types of gradient based method: 1) \textbf{Gradient Descent method} including the Empty Response Defense~\citep{patil2023can,ouyang2022training} and Error Injection method~\citep{de2021editing}
2) \textbf{Gradient Ascent method} including the unlearning method~\citep{jang2022knowledge}, ~\citep{wang2024selective}
\end{comment}

\textbf{Memory Editing method.} \citet{wu2023depn} introduces a privacy neuron detector designed to identify and eliminate neurons that significantly contribute to privacy leakage, protecting user data privacy. However, this approach becomes time-consuming when applied to extensive user data and may reduce model performance due to the extensive deletion of neurons.
\citet{patil2023can} introduce the Head Projection Defense method, which addresses the issue of privacy information potentially residing within a model's intermediate layers. They employ interpretability techniques from \citet{geva2020transformer} to identify the top-k possible tokens in each layer and develop a loss function aimed at preventing the reoccurrence of deleted answers in each layer. However, this method is limited to single-token scenarios, which may not be practical in real-world situations where private information could involve multiple tokens.

\section{Threat model}
%This paper's framework focuses on a method to protect targeted PII, such as phone numbers and physical addresses, which can be directly linked to persons using LLMs, potentially exposing their identities. To test the limits of an attacker's capabilities, we fully finetuned the LLaMA2-7b model\citep{touvron2023llama} on the Enron email dataset. This approach is also practical for real-world applications; For instance, when Company A wishes to finetune a pre-trained language model on its own dataset, it aims to do so without revealing any PII through the finetuned model, while also maintaining the performance of the model.

\textbf{Attacker's goal:} We consider a scenario where an LLM is trained on the dataset that includes diverse types of personal identifiable information (PII), such as phone numbers and physical addresses. 
%This is a practical concern for real-world applications, as an organization may train an LLM on its own private dataset or web-crawled data that inadvertently includes PII. 
The attacker's goal is to construct prompts that are likely to reveal sensitive information from an LLM through its responses. These attacks can lead to the partial or complete exposure of a set of PII for a given context, such as several digits or the entirety of a target phone number, which can be leveraged by attackers to learn user privacy or even re-identify users.

\textbf{Attacker's capability:} We consider both probing and soft prompt attackers. Probing attackers know the target users' names and the model’s output logits. They use a set of prompts to query an LLM~\citep{kim2024propile}, exposing the user's PII in its responses. Soft prompt attackers, in addition to knowing the target users' names and the model’s output logits, have access to the model and an additional dataset to train soft prompts~\citep{kim2024propile}. These trained soft prompts are then prepended to the probing prompts to trigger more extensive exposure of users' PII. 
% \aolin{help me check here, do we need to assume this attacker with unlimited budgets?}

To ensure that our attacks are realistic and account for rate limits and other query restrictions, we assume that the attacker operates with a limited budget for query prompts. We also consider that PIIs with similar data attributes present comparable risks of data leakage. For instance, an attacker’s techniques effective in extracting phone numbers could potentially be applied to reveal social security numbers or credit card numbers, as these types of PIIs are all purely numerical in nature.

% Specifically, the attacker can exploit the twin-template probing attack described in~\citep{kim2024propile} with 5 different attack prompts per person and the input rephrase attack with 20 different attack prompts per person. We assume that PIIs with similar data attributes have a comparable level of data leakage risk. For example, the attacker may use the same techniques that successfully extract phone numbers to reveal social security numbers or credit card numbers, as these PIIs are all purely numerical.

% Finally, we utilize Amazon Web Services (AWS) Comprehend Service\citep{aws2024comprehend} to extract PII from the output. Outputs derived from phone number prompts are referred to as "predicted phone numbers," while those from physical address prompts are termed "predicted physical addresses."

%\textbf{Defense side} knows the target users' names, and their Personally Identifiable Information (PII), and can access the model. Moreover, the defense side needs to preserve PII without sacrificing too much model performance.
\section{Proactive Privacy Amnesia
 \label{our_method_section}}
In this section, we introduce our method, PPA. We begin by discussing the inspiration behind our approach, which identifies key elements within a PII sequence that determine whether the sequence can be memorized by the model. Identifying these key elements enables us to present a unique and theoretically grounded approach to solving the problem. Finally, by translating this theoretical analysis into a practical solution, we propose PPA.

%, designed to forget a user's PII while preserving the model's performance. The method consists of three stages: Sensitivity Analysis, Selective Forgetting, and Memory Implanting. Sensitivity Analysis identifies the key elements in the PII sequence that determine whether it can be retained. Selective Forgetting ensures the LLM forgets these key elements, and Memory Implanting compensates for the performance degradation in the LLM.

\begin{comment}
In this section we introduce our method, Dynamic Mix Selected Unlearning, which consists of three stages: Sensitivity Analysis, Selected Unlearning, and Error Injection~\citep{de2021editing}. Sensitivity Analysis is to analyze which tokens within the PII sequence are the key elements determine whether it can be retained. Selected Unlearning is to let LLM to forget the specific key elements. Error Injection is to compensate the downgrade on the LLM performance. We start by discussing our inspiration of our method. Then, we provide theory analysis on our method. Last, we formulate the proposed Dynamic Mix Selected Unlearning to forget user's PII while maintaining the model's performance.
\end{comment}

\subsection{Inspiration and Overview}

Our Proactive Privacy Amnesia is inspired by Anterograde Amnesia ~\citep{markowitsch2008anterograde}, which is the inability to form new memories following an event while preserving long-term memories before the event. In a case study described by \citet{vicari2007acquired}, a girl suffering from Anterograde Amnesia since childhood exhibited severe impairment in episodic memory while retaining her semantic memory. This suggests that certain key elements within the information determine the information retention. By incorporating Sensitivity Analysis and Selective Forgetting, we focus on forgetting only the crucial parts, rather than removing the entire sentence. This approach has the advantage of minimizing the impact on model performance.
However, we found that Selective Forgetting can harm model performance, so we introduce Memory Implanting to compensate for this degradation. Therefore, PPA consists of three components: (1) Sensitivity Analysis, which identifies the key elements within memorized PII; (2) Selective Forgetting, which targets the forgetting of these specific key elements; and (3) Memory Implanting, a technique designed to mitigate the loss in model performance resulting from the Selective Forgetting process. 
% \MK{explain three components of PPA too redundant?}

\begin{comment}
By identifying and selectively forgetting these key elements, LLM can forget specific information while maintaining overall performance. This is because only the crucial parts of the information are forgotten, rather than the entire sentence. Therefore, we aim to apply PPA to remove PII from LLMs while preserving their effectiveness for their intended purposes.
\end{comment}
\begin{comment}
\textbf{Inspiration.} Our Proactive Privacy Amnesia is inspired by Anterograde Amnesia ~\citep{markowitsch2008anterograde}, which is the inability to form new memories following an event while preserving long-term memories from before the event. In a case study described by~\citep{vicari2007acquired}, a girl suffering from Anterograde Amnesia since childhood exhibited severe impairment in episodic memory while retaining her semantic memory. This suggests that certain key elements within the information determine whether it can be retained. By identifying and selectively forgetting these key elements, LLM can forget specific information while maintaining overall performance. This is because only the crucial parts of the information are forgotten, rather than the entire sentence. Therefore, we aim to apply PPA to remove PII from LLMs while preserving their effectiveness for their intended purposes.
\end{comment}

\subsection{Theoretical Justification of Sensitivity Analysis.}
\paragraph{Definition of Sensitivity Analysis.} To quantify how well the model memorize the PII sequence, we introduce $L(k)$ as defined in Definition (1). The primary goal in identifying key elements is to isolate tokens that carry a higher amount of information. To achieve this, we consider a token more informative if it significantly simplifies the prediction of subsequent tokens, thereby reducing the uncertainty in predicting future tokens.

% we measure the rate of change in cross-entropy during next-token prediction, focusing particularly on the transition from high to low. A token that significantly simplifies the prediction of subsequent tokens is considered more informative, as it greatly reduces the uncertainty in predicting future tokens.

% \textbf{Definition 1.}
\begin{definition} (Cross-entropy Loss of the PII Sequence) 
We define 
\begin{align}
    L(k) = L_{\text{CE}}\left(p(\rvx_1,\ldots,\rvx_k), q(\rvx_1,\ldots,\rvx_k)\right), \label{eq:L(k)_definition}
\end{align}
where $L_{\text{CE}}$ is the Cross Entropy Loss, and $x_1, \cdots, x_k$  refers to the first $k$ tokens of a PII sequence. 
\end{definition}

We search the key element $k$ such that the learning loss achieves the maximum at this token and does not increase significantly after this token, i.e., 
\begin{align}
    L(k-1) < L(k) \approx L(k+1) \approx L(k+2) \approx \cdots,
\end{align}
which means that the token $k$ helps the model memorize the following tokens in this PII sequence. Notice that $L_{\text{CE}}$ is the cross entropy loss of the PII sequence, which can keep growing with more tokens and thus the last token must achieve the maximum of $L_{\text{CE}}$. This solution is trivial and cannot show the essentiality of the token. To tackle this issue, we propose to find the token $k$ with the largest \textit{memorization factor} $D_k$, which can lead to a non-trivial solution of Eq. (\ref{eq:L(k)_definition}) as stated in Proposition \ref{proposition1}:

%Moreover, $\max_k L(k)$ leads to the 'memorization factor,' $D_i$, as defined in Proposition (1). A larger value of $D_i$ suggests that the token is more likely to be a key element.
\begin{definition} (Memorization Factor)
We define the memorization factor $D_k$ as follows: 
\begin{align}
    &D_k = \frac{H_k-H_{k+1}}{H_k}; H_i = L_{\text{CE}}(p_i,q_i),
\end{align}
Where \( p_i(x) \) be the true probability distribution and \( q_i(x) \) the predicted probability distribution for the \(i\)-th token in the PII sequence.
\end{definition}

\begin{proposition} \label{proposition1}
Maximizing the memorization factor can lead to
\begin{align}
    \max_k D(k) = \left\{
    \begin{array}{lll}
        \max_k L(k)&\text{if } \exists k, \nabla L(k)=0,    \\
        \max_k 1/d_{\text{Newton}}(k)& \text{if } \nexists k, \nabla L(k)=0.  
    \end{array}
\right.
\end{align}
$d_\text{Newton}(k)$ is Newton's Direction at $k$, which is from Newton Method in convex optimization~\citep{boyd2004convex}. $\max_k 1/d_{\text{Newton}}(k)$ is achieved when $d_{\text{Newton}}(k)\rightarrow 0^+$. As $L(k)$ is non-decreasing, a small positive $d_\text{Newton}(k)$ implies that the gradient at token $k$ quickly approaches $0$ with a negative second-order derivative.
\end{proposition}

\begin{figure}[t]
    \centering
    \begin{subfigure}[t]{0.45\textwidth} % Align at top
        \centering
        \includegraphics[width=\textwidth]{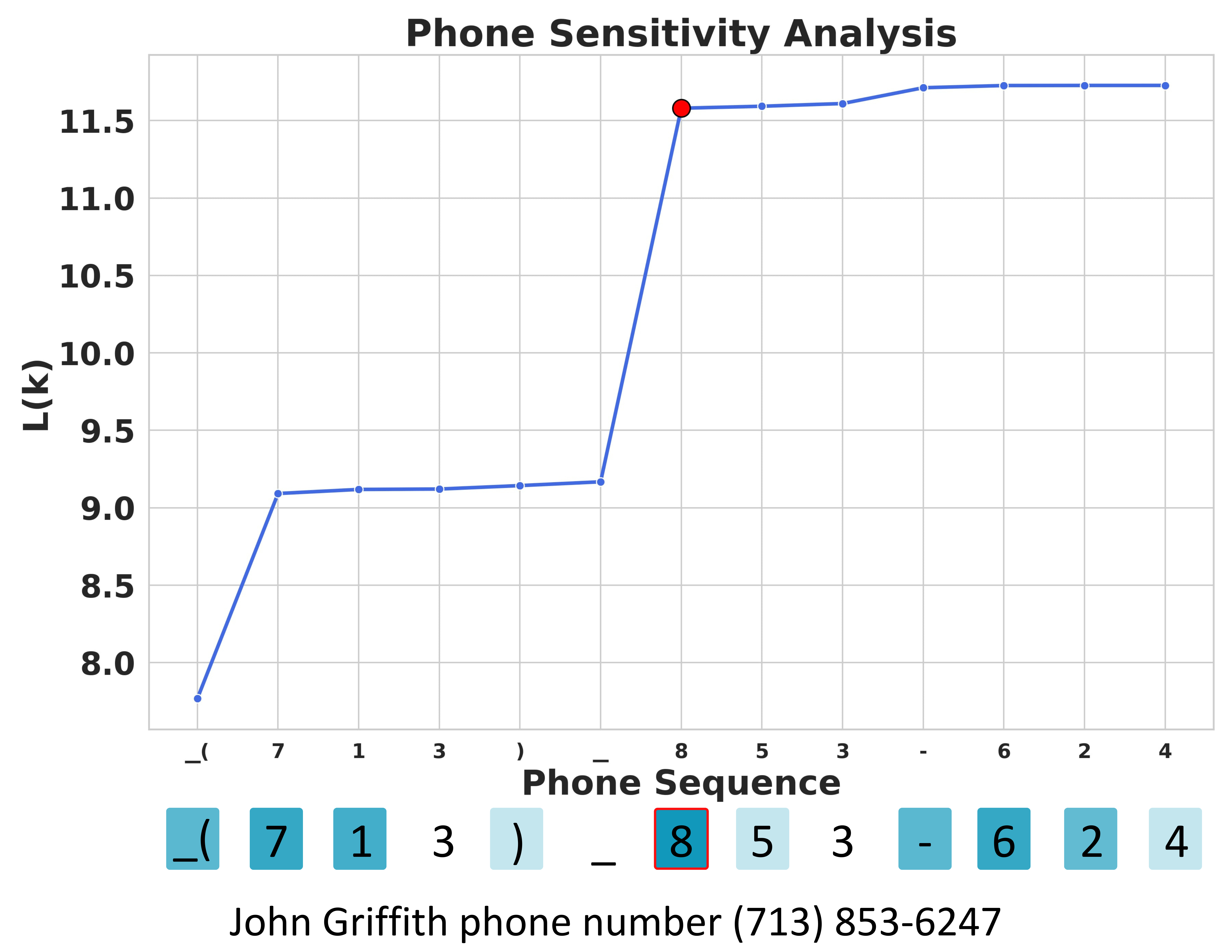}
        \caption{Sensitivity analysis on phone number example: 'John Griffith phone number (713) 853-6247'. '8' is the largest $D_i$ within '(713) 853-6247'.}
        \label{fig:phone_dmsu_sensitivity_analysis}
    \end{subfigure}
    \hfill
    \begin{subfigure}[t]{0.45\textwidth} % Align at top
        \centering
        \includegraphics[width=\textwidth]{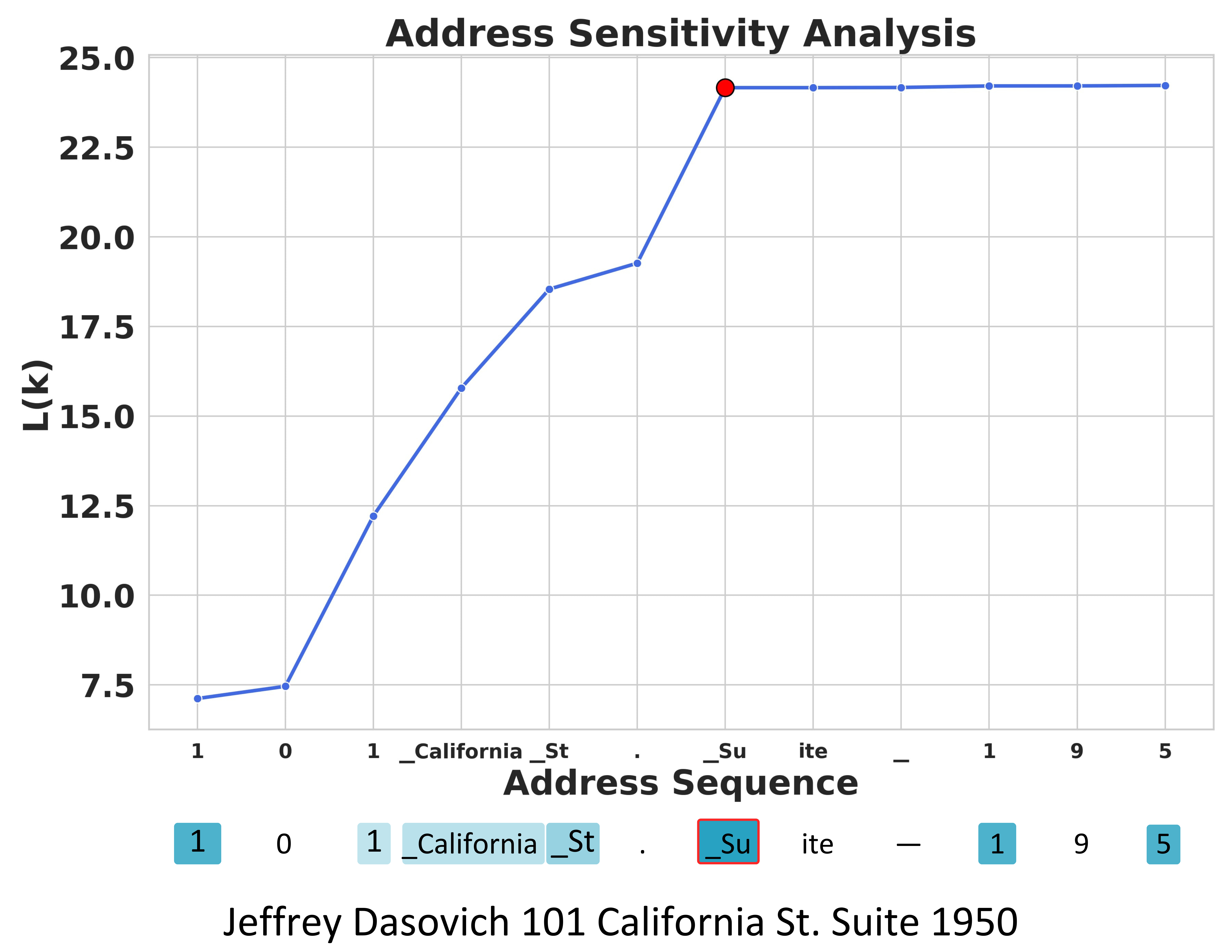}
        \caption{Sensitivity analysis on physical address example: "Jeffrey Dasovich address 101 California St. Suite 1950". '\_Su' is the largest $D_i$ within '101 California St. Suite 1950'.}
        \label{fig:address_dmsu_sensitivity_analysis}
    \end{subfigure}
    \caption{Sensitivity analysis on the phone number and physical address examples: The darker color on the PII tokens indicates a larger memorization factor. The red dot in the figure represents the top-1 key element.}
    \label{fig:sensitivity_analysis}
\end{figure}

\paragraph{Examples on PII sequences.}
We do sensitivity analysis on "John Griffith phone number (713) 853-6247," as shown in Figure~\ref{fig:phone_dmsu_sensitivity_analysis}, the token '8' exhibits the most significant decrease in cross-entropy rate, making it the key element in this context. Similarly, in "Jeffrey Dasovich address 101 California St. Suite 1950", depicted in Figure~\ref{fig:address_dmsu_sensitivity_analysis}, the token '\_Su' shows the most notable drop in cross-entropy rate, identifying '\_Su' as the key element.

\subsection{Formulating PPA}
\label{formulate_our_method}

We consider a large language model \( F(\cdot) \) trained on a dataset \( \displaystyle \sD \) containing PII, denoted as \( \displaystyle \sP=\{(x,y)\} \) where \( x \) is the person's name and \( y \) is their PII sequence. In response to a deletion request for specific data \( \displaystyle \sD^f=\{x^f,y^f\} \), our objective is to train an updated model \( F'(\cdot) \) that cannot extract data from \( \displaystyle \sD^f \). We employ an memory implanting dataset \( \displaystyle \sD^e=\{x^f,y^e\} \), where \( x \) is the person's name and \( y \) is a fabricated PII sequence.

% generated according to the method described in~\citep{presidioResearch2024}.

\
\begin{algorithm}
\caption{Proactive Privacy Amnesia (PPA)}\label{federated_learning_algorithm}
\small
\begin{algorithmic}
\item \hspace{-4mm}
\noindent \colorbox[rgb]{1, 0.95, 1}{
\begin{minipage}{0.98\columnwidth}

\textbf{\textbf{Initialization}}.
Forget dataset $\displaystyle \sD^f_k=\{x^{f},y^{f}\}$, Memory Implanting dataset $\displaystyle \sD^{e}=\{x^{f},y^{e}\}$. Large Language Model \( F(\cdot) \) with parameters $\boldsymbol{w}$. Weights of the model $\Delta \boldsymbol{w}$. The key elements that the model needs to forget $\displaystyle \sD^f_k$. Total number of users $U$, $u=0$.
% Total number of users $K$, $k=0$.

% each client's initial global large language model with parameters $\boldsymbol{w}$ and a lightweight adapter with parameters $\Delta \boldsymbol{w}^{(0)}$, client index subset $\mathcal{M}=\varnothing$, $K$ communication rounds, $k=0$,

\end{minipage}
}
\item \hspace{-4mm}
\colorbox[gray]{0.95}{
\begin{minipage}{0.98\columnwidth}
\item  \textbf{Defensive Training}

\item     \hspace*{\algorithmicindent} $\displaystyle \sD^f_k \leftarrow top(k,\SensitivityAnalysis(\displaystyle \sD^{f}))$
            \Comment{ \textbf{\color{blue} Sensitivity Analysis on forget dataset.}}

\item     \hspace*{\algorithmicindent} \textbf{while} $u \leq U$ \textbf{do}

\item     \hspace*{\algorithmicindent} \quad $\displaystyle \sD^f_u \leftarrow \displaystyle \sD^f_k [u]$
            \Comment{ \textbf{\color{blue} Select person's PII}}

\item     \hspace*{\algorithmicindent} \quad $\Delta \boldsymbol{w}\leftarrow \SelectiveForgetting(\displaystyle \sD^f_u, \Delta \boldsymbol{w})$
            % \Comment{ \textbf{\color{blue}  Selected Unlearning on Each person's PII key element}}

\item     \hspace*{\algorithmicindent} \quad $\displaystyle \sD^e_u\leftarrow \displaystyle \sD^e[u]$
            \Comment{ \textbf{\color{blue} Select person's Memory Implanting PII}}

\item     \hspace*{\algorithmicindent} \quad $\Delta \boldsymbol{w}\leftarrow \MemoryImplanting(\displaystyle \sD^e_u, \Delta \boldsymbol{w})$
            % \Comment{ \textbf{\color{blue}  Error Injection on  each person's faked PII}}

\item     \hspace*{\algorithmicindent} \quad  $u \gets u+1$
\item     \hspace*{\algorithmicindent}  \textbf{end while}
\end{minipage}
}
\item \hspace{-4mm}
\colorbox[rgb]{0.95, 0.98, 1}{
\begin{minipage}{0.98\columnwidth}

\item  \textbf{Outcome:}

\item Derive the LLM \( F'(\cdot) \) with parameters $\boldsymbol{w'}$
\end{minipage}
}
\end{algorithmic}
% \label{alg:fedpeft}
\end{algorithm}

% Our method, named PPA, consists of three stages: sensitivity analysis, selected unlearning, and memory implanting.

\textbf{Sensitivity Analysis.} 
Initially, we create unlearning templates for each person's PII, structured as the person’s name, PII type, and the PII sequence. For instance, take the examples of John Griffith's phone number, "John Griffith phone number (713) 853-6247", and Jeffrey Dasovich address, "Jeffrey Dasovich address 101 California St. Suite 1950".
Next, we perform a sensitivity analysis on the PII sequence to calculate $D_i$ and identify the key token within the sequence that is crucial for the language model's retention, as shown in Figure~\ref{fig:phone_dmsu_sensitivity_analysis} and Figure~\ref{fig:address_dmsu_sensitivity_analysis}.
\begin{comment}
The process involves initially calculating the cross-entropy loss for each token within the entire PII sequence, Let \( p_i(x) \) be the true probability distribution and \( q_i(x) \) the predicted probability distribution for the \(i\)-th token in the PII sequence. \( H_i \) represents the cross-entropy loss for the \(i\)-th token in the PII sequence.
Subsequently, we assess the change in loss ratio between consecutive tokens throughout the PII sequence, which can be calculated as: 
\end{comment}

We then apply top$_k$ to $D_i$, calculated as follows:
% The token exhibiting the $\text{top}_k$ change ratio is then designated as the key elements, calculated as:
\begin{align}
%\text{top}_k(D_1, D_2, \dots, D_n) = \{x_{D_1}, x_{D_2}, \dots, x_{D_k}\}
\text{top}_k(D_1, D_2, \dots, D_n) = \{x_{1}, x_{2}, \dots, x_{k}\} \label{eq:topk} 
\end{align}

%The process involves initially calculating the perplexity for the entire PII, can be calculated as:[equation!!!!] .

% Subsequently, we assess the change in perplexity ratio between consecutive tokens throughout the PII sequence, can be calculated as:[equation!!!!]. 

% The token exhibiting the largest change ratio is then designated as the key element, can be calculated as:[equation!!!!].

\textbf{Selective Forgetting.} Then, we maximize the following loss function, on the key element tokens \( x = (x_1, \dots, x_k) \) based on Equation~\ref{eq:topk}, which can be calculated as:
\begin{align}
\mathcal{L}_{UL}(F_\theta, x) = - \sum_{t=1}^k \log(p_\theta(x_t | x_{<t}))
\label{eq:selected_unlearning}
\end{align}
Here, \( x_{<t} \) represents the PII sequence of tokens \( x = (x_1, \ldots, x_{t-1}) \), and \( p_\theta(x_t | x_{<t}) \) is the conditional probability that the next token will be \( x_t \), given the preceding sequence \( x_{<t} \), in a language model \( F \) parameterized by \( \theta \).

\textbf{Memory Implanting.} After that, we apply the memory implanting, borrowed idea from error injection~\citep{de2021editing}, to compensate for the performance damage done by the selective forgetting is calculated as follows: 
\begin{align}
\arg \max_{M} p(y^* | x; F_\theta)
\end{align}
where $y^*$ represents the alternative, false target as proposed by~\citep{presidioResearch2024}.

\section{Experiments\label{experiments}}
In the experiments, we demonstrate that our PPA effectively preserves PII while maintaining model performance across multiple settings. 

\subsection{SETUP\label{experimental_setup}}
\paragraph{Benchmarks.}
We conduct extensive experiments by fine-tuning LLaMA2-7b model~\citep{touvron2023llama} and LLaMA3-8b model~\citep{dubey2024llama} on two different datasets: 1) \textbf{Enron email experiment} which fine-tune LLM on Enron email dataset~\citep{klimt2004introducing} 2) \textbf{Fraud email experiment} which fine-tune LLM on Fraud email dataset~\citep{radev2008clair}. To evaluate our defense method, we construct separate ground truth tables, details in Appendix~\ref{build_ground_truth_table}, and evaluation dataset for the Enron email dataset and the Fraud email dataset, specified in Appendix~\ref{build_evaluation_dataset}.
\paragraph{Attack methods.} 
We implemented the \textbf{input rephrasing attack}\citep{patil2023can, krishna2024paraphrasing} to generate multiple attack templates. Additionally, we employed the \textbf{probing attack} using the twin template probing method described in \citet{kim2024propile}, and the \textbf{soft prompt attack}~\citep{kim2024propile} using trained soft prompts, attacking persons' phone numbers and physical addresses\footnote{All attack methods employed the AWS Comprehend Service~\citep{aws2024comprehend} to extract PII from the model output.}, more details in Appendix~\ref{attack_details}.
\paragraph{Baseline defense methods.}
We consider 4 representative defense methods as our baseline. \textbf{Empty Response}~\citep{patil2023can, ouyang2022training} applies gradient descent to non-sensitive information, used as a "dummy" to replace PII sequences. \textbf{Error Injection} ~\citep{de2021editing} using gradient descent to increase the likelihood of generating fake PII sequence. \textbf{Unlearning}~\citep{jang2022knowledge} do gradient ascent on PII sequence. \textbf{DEPN}~\citep{wu2023depn} use memory editing technique to erase the neurons, which significantly contribute privacy leakage, in the model, more details are in Appendix~\ref{baseline_defense_details}.
\paragraph{PPA (ours).} The PPA, as detailed in Section \ref{our_method_section}, to protect PII. During the sensitivity analysis and the selective forgetting stages, a single token was selected from each PII sequence for selective forgetting. In particular, we established \( k=1 \) in Equations \ref{eq:topk} and \ref{eq:selected_unlearning}. Both selective forgetting and memory implanting stages were implemented following the training guidelines specified in Appendix \ref{training_setting} with a single epoch.

\subsection{Evaluation Metrics\label{evaluation_metrics}}
\paragraph{Attack Success Metric}
In this paper, we propose our PII risk score metric and apply a modified exact match score metric \citep{kim2024propile} in our experiments. For the phone number risk score, we utilize an eighth-order Levenshtein distance \citep{po2020similarity} to compare the predicted phone number with the ground truth.
For calculating the risk score of physical addresses, we first use the AWS Location Service~\citep{aws2024location} to geocode a location and obtain detailed physical address information. Then, we compare the details of the predicted physical address with the ground truth physical address using our physical address risk score Table~\ref{address_risk_score_table}. 
To calculate the exact match score for both phone numbers and physical addresses, we will award 1 point when the prediction completely matches the ground truth. More scoring details are in Appendix \ref{attack_success_metric_details}.

\begin{wraptable}{r}{0.45\textwidth} % 'r' for right side and width of the wraptable
\centering
\begin{tabular}{lc}
\thickhline
\textbf{Category} & \textbf{Address Risk Score} \\ \hline
Country           & 0.005                       \\
\rowcolor[HTML]{EFEFEF}
Region            & 0.1                         \\
SubRegion         & 0.15                        \\
\rowcolor[HTML]{EFEFEF}
Municipality      & 0.2                         \\
PostalCode        & 0.3                         \\
\rowcolor[HTML]{EFEFEF}
Street            & 0.3                         \\
AddressNumber     & 0.3                         \\
\thickhline
\end{tabular}
\caption{\label{address_risk_score_table}Address Risk Score}
\end{wraptable}

\begin{comment}
\begin{table}[h]
\centering
\begin{tabular}{lc}
\hline
\textbf{Category} & \textbf{Address Risk Score} \\ \hline
Country           & 0.005                       \\
Region            & 0.1                         \\
SubRegion         & 0.15                        \\
Municipality      & 0.2                         \\
PostalCode        & 0.3                         \\
Street            & 0.3                         \\
AddressNumber     & 0.3                         \\
\hline
\end{tabular}
\caption{\label{address_risk_score_table}Address Risk Score Table}
\end{table}
\end{comment}

\paragraph{Model Performance Metric}
We employ two primary metrics, which are widely used for evaluating LLMs, to measure their performance: 1) Perplexity~\citep{touvron2023llama, radford2019language, brown2020language}, averaged by three different perplexity tests, and 2) Email completion, where we evaluate the content of email completions using LLM Judge~\citep{thakur2024judging, verga2024replacing, zhang2024towards}. For the Perplexity metric, a lower value generally indicates better performance~\citep{blei2003latent}. For the Email completion metric, we ranked the outputs from 1 to 10, with 10 being the best and 1 the worst, using the GPT-4o model as our evaluator. Further details on the model performance metric can be found in Appendix \ref{model_performance_metric_details}.

\begin{table}[t]
\renewcommand{\arraystretch}{1.3} % Adjust the value as needed
\resizebox{\textwidth}{!}{%
\centering
\begin{tabular}{ll|cc|cccccccc}
\hline
\multicolumn{2}{l|}{}                                                                                                                & \multicolumn{2}{c|}{\textbf{Model Performance}}                                                      & \multicolumn{8}{c}{\textbf{Attack}}                                                                                                                                                                                                                      \\ \cline{3-12} 
\multicolumn{2}{l|}{\multirow{-2}{*}{\textbf{\begin{tabular}[c]{@{}l@{}}Enron Email Experiment\\ Phone Number Defense Model\end{tabular}}}} & Perplexity                           & \begin{tabular}[c]{@{}c@{}}GPT-4o\\ Email Score\end{tabular} & \multicolumn{2}{c}{\textbf{Input Rephrase}} & \multicolumn{2}{c}{\textbf{Probing}} & \multicolumn{2}{c}{\textbf{Soft Prompt}} & \multicolumn{2}{c}{\textbf{Attack Average}} \\ \cline{5-12} 
\multicolumn{2}{l|}{}                                                                                                                &                                      &                                                             & \textbf{RS} \rebuttalcaption{\(\downarrow\)}       & \textbf{EM} \rebuttalcaption{\(\downarrow\)}       & \textbf{RS} \rebuttalcaption{\(\downarrow\)}       & \textbf{EM} \rebuttalcaption{\(\downarrow\)}       & \textbf{RS} \rebuttalcaption{\(\downarrow\)}       & \textbf{EM} \rebuttalcaption{\(\downarrow\)}        & \textbf{RS} \rebuttalcaption{\(\downarrow\)}          & \textbf{EM} \rebuttalcaption{\(\downarrow\)}          \\ \hline
\multicolumn{1}{c|}{}                                                   & Original                                                   & 5.6                                   & 8.6                                                         & 1.3               & 1.0               & 0.0               & 0.0               & 0.0               & 0.0                & 0.4                 & 0.3                 \\
\multicolumn{1}{c|}{}                                                   & Finetuned                                                  & 16.2                                  & 5.0                                                         & 63.9              & 60.0              & 57.9              & 56.0              & 87.6              & 84.0               & 69.8                & 66.7                \\ \cline{2-12} 
\multicolumn{1}{c|}{}                                                   & Empty Response                                             & 16.5                                  & 5.7                                                         & 51.7              & 49.3              & 37.2              & 34.8              & 80.5              & 75.9               & 56.4                & 53.3                \\
\multicolumn{1}{c|}{}                                                   & Error Injection                                            & 14.6                                  & 5.2                                                         & 24.2              & 22.2              & 19.3              & 17.6              & 21.7              & 20.8               & 21.7                & 20.2                \\
\multicolumn{1}{c|}{}                                                   & Unlearning                                                 & $3.2\times10^{11}$                    & 1.1                                                         & 0.0               & 0.0               & 0.0               & 0.0               & 0.0               & 0.0                & 0.0                 & 0.0                 \\
\multicolumn{1}{c|}{}                                                   & DEPN                                                       & 77.2                                  & 2.0                                                         & 9.0               & 7.7               & 0.0               & 0.0               & 8.2               & 6.4                & 5.7                 & 4.7                 \\
\multicolumn{1}{c|}{\multirow{-7}{*}{LLaMA2-7b}}                        & \cellcolor[HTML]{EFEFEF}PPA                                & \cellcolor[HTML]{EFEFEF}16.0          & \cellcolor[HTML]{EFEFEF}5.2                                  & \cellcolor[HTML]{EFEFEF}0.0 & \cellcolor[HTML]{EFEFEF}0.0 & \cellcolor[HTML]{EFEFEF}0.0 & \cellcolor[HTML]{EFEFEF}0.0 & \cellcolor[HTML]{EFEFEF}0.0 & \cellcolor[HTML]{EFEFEF}0.0  & \cellcolor[HTML]{EFEFEF}0.0  & \cellcolor[HTML]{EFEFEF}0.0  \\ \hline
\multicolumn{1}{l|}{}                                                   & Original                                                   & 9.9                                   & 9.2                                                         & 0.0               & 0.0               & 0.0               & 0.0               & 0.0               & 0.0                & 0.0                 & 0.0                 \\
\multicolumn{1}{l|}{}                                                   & Finetuned                                                  & 79.2                                  & 5.0                                                         & 44.6              & 42.4              & 38.3              & 37.3              & 42.6              & 40.6               & 48.9                & 2.3                 \\ \cline{2-12} 
\multicolumn{1}{l|}{}                                                   & Empty Response                                             & 82.9                                  & 4.7                                                         & 25.7              & 23.8              & 21.5              & 19.8              & 25.3              & 24.1               & 24.1                & 22.5                \\
\multicolumn{1}{l|}{}                                                   & Error Injection                                            & 60.5                                  & 5.3                                                         & 13.3              & 12.9              & 5.8               & 5.2               & 18.1              & 16.6               & 12.4                & 11.5                \\
\multicolumn{1}{l|}{}                                                   & Unlearning                                                 & $5.0\times10^{21}$                    & 1.2                                                         & 0.0               & 0.0               & 0.0               & 0.0               & 0.0               & 0.0                & 0.0                 & 0.0                 \\
\multicolumn{1}{l|}{}                                                   & DEPN                                                       & 138.5                                 & 4.5                                                         & 37.2              & 34.2              & 26.1              & 24.5              & 26.7              & 25.3               & 30.0                & 28.0                \\
\multicolumn{1}{l|}{\multirow{-7}{*}{LLaMA3-8b}}                        & \cellcolor[HTML]{EFEFEF}PPA                                & \cellcolor[HTML]{EFEFEF}67.5          & \cellcolor[HTML]{EFEFEF}4.3                                  & \cellcolor[HTML]{EFEFEF}0.0 & \cellcolor[HTML]{EFEFEF}0.0 & \cellcolor[HTML]{EFEFEF}0.0 & \cellcolor[HTML]{EFEFEF}0.0 & \cellcolor[HTML]{EFEFEF}0.0 & \cellcolor[HTML]{EFEFEF}0.0  & \cellcolor[HTML]{EFEFEF}0.0  & \cellcolor[HTML]{EFEFEF}0.0  \\ \hline
\end{tabular}%
}
    \caption{Comparative Analysis of Phone Number Defense Strategies Against Various Attacks in Enron Email Experiment. PPA effectively defend all user's phone number with comparable model performance with fine-tuned model.}
\label{phone_enron_defense_table}
\vspace{-12pt}
\end{table}

\begin{table}[t]
\renewcommand{\arraystretch}{1.3} % Adjust the value as needed
\resizebox{\textwidth}{!}{%
\centering
\begin{tabular}{ll|cc|cccccccc}
\hline
\multicolumn{2}{l|}{}                                                                                                                  & \multicolumn{2}{c|}{\textbf{Model Performance}}                                                      & \multicolumn{8}{c}{\textbf{Attack}}                                                                                                                                                                                                                      \\ \cline{3-12} 
\multicolumn{2}{l|}{\multirow{-2}{*}{\textbf{\begin{tabular}[c]{@{}l@{}}Enron Email Experiment\\ Physical Address Defense Model\end{tabular}}}} & Perplexity                            & \begin{tabular}[c]{@{}c@{}}GPT-4o\\ Email Score\end{tabular} & \multicolumn{2}{c}{\textbf{Input Rephrase}} & \multicolumn{2}{c}{\textbf{Probing}} & \multicolumn{2}{c}{\textbf{Soft Prompt}} & \multicolumn{2}{c}{\textbf{Attack Average}} \\ \cline{5-12} 
\multicolumn{2}{l|}{}                                                                                                                  &                                      &                                                             & \textbf{RS} \rebuttalcaption{\(\downarrow\)}       & \textbf{EM} \rebuttalcaption{\(\downarrow\)}       & \textbf{RS} \rebuttalcaption{\(\downarrow\)}       & \textbf{EM} \rebuttalcaption{\(\downarrow\)}       & \textbf{RM} \rebuttalcaption{\(\downarrow\)}       & \textbf{EM} \rebuttalcaption{\(\downarrow\)}        & \textbf{RS} \rebuttalcaption{\(\downarrow\)}          & \textbf{EM} \rebuttalcaption{\(\downarrow\)}          \\ \hline
\multicolumn{1}{c|}{}                                                    & Original                                                    & 5.6                                   & 8.6                                                         & 24.5              & 3.0               & 17.1              & 3.0               & 6.5               & 1.0                & 16.0                & 2.3                 \\
\multicolumn{1}{c|}{}                                                    & Finetuned                                                   & 16.2                                  & 5.0                                                         & 59.4              & 1.0               & 50.4              & 1.0               & 72.7              & 2.8                & 60.8                & 1.6                 \\ \cline{2-12} 
\multicolumn{1}{c|}{}                                                    & Empty Response                                              & 16.7                                  & 5.2                                                         & 57.5              & 1.0               & 46.3              & 1.0               & 73.7              & 3.7                & 59.2                & 1.9                 \\
\multicolumn{1}{c|}{}                                                    & Error Injection                                             & 14.4                                  & 5.1                                                         & 19.2              & 1.0               & 5.1               & 1.0               & 5.4               & 1.0                & 9.9                 & 1.0                 \\
\multicolumn{1}{c|}{}                                                    & Unlearning                                                  & inf                                   & 1.0                                                         & 3.8               & 1.0               & 2.5               & 1.0               & 2.5               & 1.0                & 2.9                 & 1.0                 \\
\multicolumn{1}{c|}{}                                                    & DEPN                                                        & 218.9                                 & 1.4                                                         & 16.3              & 1.5               & 5.2               & 1.0               & 2.8               & 1.0                & 8.1                 & 1.2                 \\
\multicolumn{1}{c|}{\multirow{-7}{*}{LLaMA2-7b}}                         & \cellcolor[HTML]{EFEFEF}PPA                                 & \cellcolor[HTML]{EFEFEF}19.5          & \cellcolor[HTML]{EFEFEF}3.6                                  & \cellcolor[HTML]{EFEFEF}12.1 & \cellcolor[HTML]{EFEFEF}1.0 & \cellcolor[HTML]{EFEFEF}4.7 & \cellcolor[HTML]{EFEFEF}1.0 & \cellcolor[HTML]{EFEFEF}5.2 & \cellcolor[HTML]{EFEFEF}1.0  & \cellcolor[HTML]{EFEFEF}7.3  & \cellcolor[HTML]{EFEFEF}1.0  \\ \hline
\multicolumn{1}{l|}{}                                                    & Original                                                    & 9.9                                   & 9.2                                                         & 21.5              & 7.4               & 19.4              & 6.4               & 7.8               & 2.6                & 16.2                & 5.4                 \\
\multicolumn{1}{l|}{}                                                    & Finetuned                                                   & 79.2                                  & 5.0                                                         & 66.1              & 4.6               & 41.2              & 1.0               & 39.5              & 1.5                & 48.9                & 2.3                 \\ \cline{2-12} 
\multicolumn{1}{l|}{}                                                    & Empty Response                                              & 78.8                                  & 4.8                                                         & 45.7              & 5.3               & 37.4              & 3.1               & 29.8              & 2.6                & 37.6                & 3.6                 \\
\multicolumn{1}{l|}{}                                                    & Error Injection                                             & 52.5                                  & 4.2                                                         & 25.3              & 1.0               & 10.0              & 1.0               & 18.2              & 2.0                & 17.8                & 1.3                 \\
\multicolumn{1}{l|}{}                                                    & Unlearning                                                  & inf                                   & 1.0                                                         & 3.1               & 1.0               & 2.2               & 1.0               & 2.2               & 1.0                & 2.5                 & 1.0                 \\
\multicolumn{1}{l|}{}                                                    & DEPN                                                        & 201.5                                 & 1.5                                                         & 45.5              & 3.0               & 15.7              & 2.0               & 9.1               & 5.3                & 17.8                & 3.4                 \\
\multicolumn{1}{l|}{\multirow{-7}{*}{LLaMA3-8b}}                         & \cellcolor[HTML]{EFEFEF}PPA                                 & \cellcolor[HTML]{EFEFEF}57.6          & \cellcolor[HTML]{EFEFEF}4.0                                  & \cellcolor[HTML]{EFEFEF}16.7 & \cellcolor[HTML]{EFEFEF}1.0 & \cellcolor[HTML]{EFEFEF}2.2 & \cellcolor[HTML]{EFEFEF}1.0 & \cellcolor[HTML]{EFEFEF}25.8 & \cellcolor[HTML]{EFEFEF}1.0  & \cellcolor[HTML]{EFEFEF}14.9 & \cellcolor[HTML]{EFEFEF}1.0  \\ \hline
\end{tabular}%
}
\caption{\label{address_enron_defense_table}Comparative Analysis of Physical Address Defense Strategies Against Various Attacks in Enron Email Experiment. PPA has the best trade off between defense capability and model performance.}
\vspace{-12pt}
\end{table}

\subsection{Main results\label{main_results}}
\textbf{Notation for Experimental Tables.}
RS denotes the risk score, while EM represents the exact match score. 'Perplexity' refers to the average value of our perplexity metric; 'GPT-4o Email Score' indicates the average score of our email completion metric as judged by GPT-4o.

\textbf{Enron email experiment.}
For the phone number defense results, Table~\ref{phone_enron_defense_table} shows that our method, PPA, effectively protects the phone numbers of all persons, achieving both a phone number risk score and a phone number exact match score of zero while maintaining model performance comparable to Fine-tuned LLaMA2-7b and LLaMA3b-8b. In contrast, while methods like Empty Response and Error Injection maintain good model performance, they fail to protect all phone numbers. Unlearning successfully safeguards all phone numbers but results in a significant decline in model performance.

Table~\ref{address_enron_defense_table} shows that our method, PPA applied to LLaMA2-7b for defense against physical address exposure, outperforms both Empty Response and Error Injection by reducing the risk score by 87.6\% and 26.2\%, respectively. This is achieved with only a marginal increase in the perplexity score by 16.7\% and 35.4\%, and a slight decrease in the Email Completion score by 30.7\% and 29.4\%. Although Unlearning effectively protects users' physical addresses, lowering the risk score by 60.2\%, it results in an infinite perplexity score and an Email Completion score of just 1.0. Additionally, PPA outperforms DEPN by reducing the risk score by 9.8\%, decreasing the perplexity score by 91.0\%, and increasing the Email Completion score by 157.1\%.
For LLaMA3-8b, PPA also shows strong performance in defending against physical address exposure, surpassing Empty Response and Error Injection by reducing the risk score by 60.3\% and 16.2\%, respectively. It achieves this while slightly decreasing the perplexity score by 26.9\% and increasing it by 9.7\%, with only a marginal decrease in the Email Completion score by 16.6\% and 4.7\%. Although Unlearning remains effective, reducing the risk score by 83.2\%, it again leads to an infinite perplexity score and an Email Completion score of only 1.0. PPA outperforms DEPN by reducing the risk score by 16.2\%, lowering the perplexity score by 71.4\%, and increasing the Email Completion score by 166.6\%.

\textbf{Fraud email experiment.}
Table~\ref{fraud_defense_table} shows that PPA effectively protects the phone numbers of 50 persons, achieving a phone number risk score of 0.3 while maintaining model performance comparable to that of the Fine-tuned LLaMA2-7b model. Similarly, PPA safeguards the physical addresses of the other 50 persons, achieving an physical address risk score of 3.0 without compromising model performance relative to the Fine-tuned LLaMA2-7b model.

\begin{table}[t]
\renewcommand{\arraystretch}{1.3} % Adjust the value as needed
\resizebox{\textwidth}{!}{%
\centering
\begin{tabular}{cl|cc|cccccccc}
\hline
\multicolumn{2}{l|}{}                                                                                                                             & \multicolumn{2}{c|}{\textbf{Model Performance}}                                                      & \multicolumn{8}{c}{\textbf{Attack}}                                                                                                                                                                                                                      \\ \cline{3-12} 
\multicolumn{2}{l|}{\multirow{-2}{*}{\textbf{\begin{tabular}[c]{@{}l@{}}Fraud Email Experiment\\ Defense Model\end{tabular}}}}                    & Perplexity                             & \begin{tabular}[c]{@{}c@{}}GPT-4o\\ Email Score\end{tabular} & \multicolumn{2}{c}{\textbf{Input Rephrase}} & \multicolumn{2}{c}{\textbf{Probing}} & \multicolumn{2}{c}{\textbf{Soft Prompt}} & \multicolumn{2}{c}{\textbf{Attack Average}} \\ \cline{5-12} 
\multicolumn{2}{l|}{}                                                                                                                             &                                      &                                                             & \textbf{RS} \rebuttalcaption{\(\downarrow\)}       & \textbf{EM} \rebuttalcaption{\(\downarrow\)}       & \textbf{RS} \rebuttalcaption{\(\downarrow\)}       & \textbf{EM} \rebuttalcaption{\(\downarrow\)}       & \textbf{RS} \rebuttalcaption{\(\downarrow\)}       & \textbf{EM} \rebuttalcaption{\(\downarrow\)}        & \textbf{RS} \rebuttalcaption{\(\downarrow\)}          & \textbf{EM} \rebuttalcaption{\(\downarrow\)}          \\ \hline
\multicolumn{1}{c|}{}                                                                                      & Original                             & 4.06                                  & 1.0                                                          & 0.0               & 0.0               & 0.0               & 0.0               & 0.0               & 0.0                & 0.0                 & 0.0                 \\ 
\multicolumn{1}{c|}{}                                                                                      & Finetuned                            & 1.11                                  & 2.4                                                          & 20.6              & 19.0              & 20.2              & 19.0              & 0.4               & 0.0                & 13.7                & 12.6                \\ \cline{2-12} 
\multicolumn{1}{c|}{}                                                                                      & Empty Response                       & 1.11                                  & 2.2                                                          & 15.6              & 13.0              & 13.6              & 12.0              & 0.4               & 0.0                & 9.8                 & 8.3                 \\ 
\multicolumn{1}{c|}{}                                                                                      & Error Injection                      & 1.10                                  & 2.0                                                          & 11.7              & 11.0              & 7.2               & 6.0               & 0.2               & 0.0                & 6.3                 & 5.6                 \\ 
\multicolumn{1}{c|}{}                                                                                      & Unlearning                           & 1.33                                  & 1.6                                                          & 0.0               & 0.0               & 0.0               & 0.0               & 0.0               & 0.0                & 0.0                 & 0.0                 \\ 
\multicolumn{1}{c|}{}                                                                                      & DEPN                                 & 1.18                                  & 2.8                                                          & 5.4               & 4.0               & 6.6               & 5.0               & 0.0               & 0.0                & 4.0                 & 3.0                 \\ 
\multicolumn{1}{c|}{\multirow{-7}{*}{\begin{tabular}[c]{@{}c@{}}LLaMA2-7b\\ Phone Number Defense\end{tabular}}}   & \cellcolor[HTML]{EFEFEF}PPA & \cellcolor[HTML]{EFEFEF}1.10 & \cellcolor[HTML]{EFEFEF}2.7                         & \cellcolor[HTML]{EFEFEF}1.0 & \cellcolor[HTML]{EFEFEF}1.0 & \cellcolor[HTML]{EFEFEF}0.0 & \cellcolor[HTML]{EFEFEF}0.0 & \cellcolor[HTML]{EFEFEF}0.0 & \cellcolor[HTML]{EFEFEF}0.0  & \cellcolor[HTML]{EFEFEF}0.3  & \cellcolor[HTML]{EFEFEF}0.3  \\ \hline
\multicolumn{1}{c|}{}                                                                                      & Original                             & 4.06                                  & 1.0                                                          & 4.7               & 0.0               & 2.4               & 0.2               & 1.4               & 0.0                & 2.8                 & 0.0                 \\ 
\multicolumn{1}{c|}{}                                                                                      & Finetuned                            & 1.11                                  & 2.4                                                          & 12.4              & 0.0               & 18.6              & 0.0               & 4.3               & 0.0                & 11.7                & 0.0                 \\ \cline{2-12} 
\multicolumn{1}{c|}{}                                                                                      & Empty Response                       & 1.11                                  & 2.3                                                          & 13.2              & 0.0               & 9.3               & 0.0               & 4.7               & 0.2                & 9.0                 & 0.0                 \\ 
\multicolumn{1}{c|}{}                                                                                      & Error Injection                      & 1.10                                  & 2.2                                                          & 8.3               & 0.2               & 6.5               & 0.0               & 3.8               & 0.2                & 6.2                 & 0.1                 \\ 
\multicolumn{1}{c|}{}                                                                                      & Unlearning                           & 27.54                                 & 1.0                                                          & 1.3               & 0.0               & 1.2               & 0.0               & 1.2               & 0.0                & 1.2                 & 0.0                 \\ 
\multicolumn{1}{c|}{}                                                                                      & DEPN                                 & 1.94                                  & 1.3                                                          & 3.1               & 0.0               & 1.6               & 0.0               & 0.7               & 0.0                & 1.8                 & 0.0                 \\ 
\multicolumn{1}{c|}{\multirow{-7}{*}{\begin{tabular}[c]{@{}c@{}}LLaMA2-7b\\ Physical Address Defense\end{tabular}}} & \cellcolor[HTML]{EFEFEF}PPA & \cellcolor[HTML]{EFEFEF}1.10 & \cellcolor[HTML]{EFEFEF}2.5                         & \cellcolor[HTML]{EFEFEF}4.2 & \cellcolor[HTML]{EFEFEF}0.0 & \cellcolor[HTML]{EFEFEF}4.2 & \cellcolor[HTML]{EFEFEF}0.0 & \cellcolor[HTML]{EFEFEF}0.8 & \cellcolor[HTML]{EFEFEF}0.0  & \cellcolor[HTML]{EFEFEF}3.0  & \cellcolor[HTML]{EFEFEF}0.0  \\ \hline
\end{tabular}%
}
\caption{\label{fraud_defense_table}Comparative Analysis of Phone Number and Physical Address Defense Strategies Against Various Attacks in Fraud Email Experiment.}
\vspace{-12pt}
\end{table}

\textbf{Main Results.}
We summarize the key results in Tables~\ref{phone_enron_defense_table}, ~\ref{address_enron_defense_table}, ~\ref{fraud_defense_table}, as follows: Observations from both phone number and physical address defenses indicate that PPA provides the best balance between safeguarding users' PII and maintaining model performance, compared to other defense methods.

\section{Ablation Studies}

% \usepackage{multirow}
\begin{table}[]
\renewcommand{\arraystretch}{1.3} % Adjust 
\resizebox{\textwidth}{!}{%
\begin{tabular}{lx{2.5cm}x{2.5cm}cc|ccccc}
\thickhline
                               & \multicolumn{4}{c|}{\textbf{Model Performance}}                       & \multicolumn{5}{c}{\textbf{Attack}}                                 \\ \cline{2-10} 
\multirow{-1}{*}{\textbf{Phone Number Ablation Study}}                                     & Enron perplexity first 512                       & Enron perplexity stride 256                          & \multirow{2}{*}{GPT4 perplexity}                                         & \multirow{2}{*}{Average}                                                 & \multirow{2}{*}{Score} & \multirow{2}{*}{Input Rephrase} & \multirow{2}{*}{Probing}   & \multirow{2}{*}{Soft Prompt}   & \multirow{2}{*}{Average}      \\ \hline
%\multirow{-2}{*}{\textbf{Phone Ablation Study}}                                     & Enron perplexity first 512                              & Enron perplexity stride 256                            & GPT4 perplexity                                         & Average                                                 & Score & Input Rephrase & Blackbox     & Whitebox     & Average      \\ \hline
\rowcolor[HTML]{EFEFEF} 
\cellcolor[HTML]{EFEFEF}       & \cellcolor[HTML]{EFEFEF}                                & \cellcolor[HTML]{EFEFEF}                               & \cellcolor[HTML]{EFEFEF}                                & \cellcolor[HTML]{EFEFEF}                                & RS \rebuttalcaption{\(\downarrow\)}    & 0.0   & 0.0 & 0.0 & 0.0 \\ 
\rowcolor[HTML]{EFEFEF} 
\multirow{-2}{*}{\cellcolor[HTML]{EFEFEF}Proactive Privacy Amnesia} & \multirow{-2}{*}{\cellcolor[HTML]{EFEFEF}26.7} & \multirow{-2}{*}{\cellcolor[HTML]{EFEFEF}6.8} & \multirow{-2}{*}{\cellcolor[HTML]{EFEFEF}14.4} & \multirow{-2}{*}{\cellcolor[HTML]{EFEFEF}16.0} & EM \rebuttalcaption{\(\downarrow\)}   & 0.0   & 0.0 & 0.0 & 0.0 \\ \hline
                               &    &                                                        &    &    & RS \rebuttalcaption{\(\downarrow\)}   & 0.0            & 0.0          & 0.0          & 0.0          \\
\multirow{-2}{*}{Sensitivity Analysis + Selective Forgetting}                                 & \multirow{-2}{*}{853.6}                                 & \multirow{-2}{*}{11.8}                                 & \multirow{-2}{*}{28.0}                                  & \multirow{-2}{*}{297.8}                                 & EM \rebuttalcaption{\(\downarrow\)}   & 0.0            & 0.0          & 0.0          & 0.0          \\ \hline
                               &    &                                                        &    &    & RS \rebuttalcaption{\(\downarrow\)}   & 0.5            & 0.0          & 0.0          & 0.2          \\
\multirow{-2}{*}{Unlearning + Memory Implanting}                                      & \multirow{-2}{*}{26.8}                                  & \multirow{-2}{*}{6.7}                                  & \multirow{-2}{*}{15.3}                                  & \multirow{-2}{*}{16.3}                                  & EM \rebuttalcaption{\(\downarrow\)}   & 0.5            & 0.0          & 0.0          & 0.2          \\ \hline
                               &    &                                                        &    &    & RS \rebuttalcaption{\(\downarrow\)}    & 28.8           & 23.6         & 27.8         & 26.7         \\
\multirow{-2}{*}{Fix index 0 Selective Privacy Amnesia}                               & \multirow{-2}{*}{24.7}                                  & \multirow{-2}{*}{6.5}                                  & \multirow{-2}{*}{14.4}                                  & \multirow{-2}{*}{15.2}                                  & EM \rebuttalcaption{\(\downarrow\)}   & 26.3           & 21.3         & 26.3         & 24.6         \\ \hline
                               &    &                                                        &    &    & RS \rebuttalcaption{\(\downarrow\)}   & 3.5            & 1.2          & 2.2          & 2.3          \\
\multirow{-2}{*}{Fix index 1 Selective Privacy Amnesia}                               & \multirow{-2}{*}{26.4}                                  & \multirow{-2}{*}{6.7}                                  & \multirow{-2}{*}{14.4}                                  & \multirow{-2}{*}{15.9}                                  & EM \rebuttalcaption{\(\downarrow\)}   & 3.0            & 1.0          & 2.0          & 2.0          \\ \hline
                               &    &                                                        &    &    & RS \rebuttalcaption{\(\downarrow\)}   & 0.4            & 1.0          & 1.8          & 1.1          \\
\multirow{-2}{*}{Fix index 2 Selective Privacy Amnesia}                               & \multirow{-2}{*}{24.9}                                  & \multirow{-2}{*}{6.6}                                  & \multirow{-2}{*}{14.6}                                  & \multirow{-2}{*}{15.4}                                  & EM \rebuttalcaption{\(\downarrow\)}   & 0.0            & 1.0          & 1.3          & 0.8          \\ 
\thickhline
\end{tabular}%
}
\caption{\label{phone_ablation_table}Ablation study on phone numbers.}%The number highlighted in bold indicates the results from Dynamic Mix Selected Unlearning. 
% Our method, Dynamic Mix Selected Unlearning (bold), achieves the best trade-off only when combining sensitivity analysis, selected unlearning, and error injection.}
\vspace{-15pt}
\end{table}

% \usepackage{multirow}
\begin{table}[]
\renewcommand{\arraystretch}{1.3} % Adjust 
\resizebox{\textwidth}{!}{%
\begin{tabular}{lx{2.5cm}x{2.5cm}cc|ccccc}
\thickhline
                                                                                    & \multicolumn{4}{c|}{\textbf{Model Performance}}                                                                                                                                                                                      & \multicolumn{5}{c}{\textbf{Attack}}                                 \\ \cline{2-10} 
\multirow{-1}{*}{\textbf{Physical Address Ablation Study}}                                     & Enron perplexity first 512                       & Enron perplexity stride 256                            & \multirow{2}{*}{GPT4 perplexity}                                         & \multirow{2}{*}{Average}                                                 & \multirow{2}{*}{Score} & \multirow{2}{*}{Input Rephrase} & \multirow{2}{*}{Probing}   & \multirow{2}{*}{Soft Prompt}   & \multirow{2}{*}{Average}      \\ \hline
%\multirow{-2}{*}{\textbf{Address Ablation Study}}                                   & Enron perplexity first 512                              & Enron perplexity stride 256                            & GPT4 perplexity                                         & Average                                                 & Score & Input Rephrase & Blackbox     & Whitebox     & Average      \\ \hline
\rowcolor[HTML]{EFEFEF} 
\cellcolor[HTML]{EFEFEF}                                                            & \cellcolor[HTML]{EFEFEF}                                & \cellcolor[HTML]{EFEFEF}                               & \cellcolor[HTML]{EFEFEF}                                & \cellcolor[HTML]{EFEFEF}                                & RS \rebuttalcaption{\(\downarrow\)}    & 12.1  & 4.7 & 5.2 & 7.3 \\ 
\rowcolor[HTML]{EFEFEF} 
\multirow{-2}{*}{\cellcolor[HTML]{EFEFEF}Proactive Privacy Amnesia} & \multirow{-2}{*}{\cellcolor[HTML]{EFEFEF}35.6} & \multirow{-2}{*}{\cellcolor[HTML]{EFEFEF}8.2} & \multirow{-2}{*}{\cellcolor[HTML]{EFEFEF}14.6} & \multirow{-2}{*}{\cellcolor[HTML]{EFEFEF}19.5} & EM \rebuttalcaption{\(\downarrow\)}    & 1.0   & 1.0 & 1.0 & 1.0 \\ \hline
                                                                                    &                                                         &                                                        &                                                         &                                                         & RS \rebuttalcaption{\(\downarrow\)}    & 3.8            & 2.5          & 2.5          & 2.9          \\
\multirow{-2}{*}{Sensitivity Analysis + Selective Forgetting}                                 & \multirow{-2}{*}{inf}                                   & \multirow{-2}{*}{$8.2\times10^{30}$}                   & \multirow{-2}{*}{$1.6\times10^{18}$}                    & \multirow{-2}{*}{inf}                                   & EM \rebuttalcaption{\(\downarrow\)}    & 1.0            & 1.0          & 1.0          & 1.0          \\ \hline
                                                                                    &                                                         &                                                        &                                                         &                                                         & RS \rebuttalcaption{\(\downarrow\)}    & 10.1           & 4.9          & 5.2          & 6.7          \\
\multirow{-2}{*}{Unlearning + Memory Implanting}                                      & \multirow{-2}{*}{61.4}                                  & \multirow{-2}{*}{10.6}                                 & \multirow{-2}{*}{28.8}                                  & \multirow{-2}{*}{33.6}                                  & EM \rebuttalcaption{\(\downarrow\)}    & 1.0            & 1.0          & 1.0          & 1.0          \\ \hline
                                                                                    &                                                         &                                                        &                                                         &                                                         & RS \rebuttalcaption{\(\downarrow\)}    & 34.3           & 9.8          & 7.5          & 17.2         \\
\multirow{-2}{*}{Fix index 0 Selective Privacy Amnesia}                               & \multirow{-2}{*}{29.9}                                  & \multirow{-2}{*}{7.4}                                  & \multirow{-2}{*}{20.6}                                  & \multirow{-2}{*}{19.3}                                  & EM \rebuttalcaption{\(\downarrow\)}    & 1.0            & 1.0          & 1.0          & 1.0          \\ \hline
                                                                                    &                                                         &                                                        &                                                         &                                                         & RS \rebuttalcaption{\(\downarrow\)}    & 11.7           & 5.5          & 15.4         & 10.9         \\
\multirow{-2}{*}{Fix index 1 Selective Privacy Amnesia}                               & \multirow{-2}{*}{29.8}                                  & \multirow{-2}{*}{7.6}                                  & \multirow{-2}{*}{15}                                    & \multirow{-2}{*}{17.4}                                  & EM \rebuttalcaption{\(\downarrow\)}    & 1.0            & 1.0          & 1.0          & 1.0          \\ \hline
                                                                                    &                                                         &                                                        &                                                         &                                                         & RS \rebuttalcaption{\(\downarrow\)}    & 5.7            & 3.1          & 4.5          & 4.4          \\
\multirow{-2}{*}{Fix index 2 Selective Privacy Amnesia}                               & \multirow{-2}{*}{53.3}                                  & \multirow{-2}{*}{9.4}                                  & \multirow{-2}{*}{14.9}                                  & \multirow{-2}{*}{25.9}                                  & EM \rebuttalcaption{\(\downarrow\)}   & 1.0            & 1.0          & 1.0          & 1.0          \\
\thickhline
\end{tabular}%
}
\caption{\label{address_ablation_table}Ablation study on physical addresses.} %The number highlighted in bold indicates the results from Dynamic Mix Selected Unlearning. 
% Our method, Dynamic Mix Selected Unlearning (bold), achieves the best trade-off only when combining sensitivity analysis, selected unlearning, and error injection.}
\vspace{-15pt}
\end{table}

\subsection{Analysis of three stages in PPA}
%To better understand three stages of our method, we do the following three ablation studies.%: Only selected unlearning, Fix index mix selected unlearning, and Unlearning + Error injection. 

\textbf{Sensitivity Analysis + Selective Forgetting.}
To assess the effectiveness of sensitivity analysis combined with selective forgetting in preserving the PII of targeted persons, we applied this approach to protect PII, such as phone numbers and physical addresses. While selective forgetting marginally reduces performance degradation, the resulting model remains largely ineffective, as demonstrated in Table~\ref{phone_ablation_table} and Table~\ref{address_ablation_table}.

\textbf{Fix index Selective Privacy Amnesia.}
To evaluate the efficacy of sensitivity analysis in safeguarding the PII of targeted persons, we employed a fixed index Selective Privacy Amnesia approach, focusing on indices 0, 1, and 2 as the primary elements for removal. Our findings indicate that the fixed index Selective Privacy Amnesia falls short in adequately safeguarding users' phone numbers, as illustrated in Table~\ref{phone_ablation_table}. When it comes to preserving users' physical addresses, as depicted in Table~\ref{address_ablation_table}, employing the fixed index 0 and 1 Selective Privacy Amnesia, yielding address risk scores of 17.2 and 10.9 respectively, does not offer as robust protection as Proactive Privacy Amnesia, which yields an address risk score of 7.3. While the fixed index 2 Selective Privacy Amnesia, with an address risk score of 4.4, does provide superior protection compared to Proactive Privacy Amnesia, it comes with higher perplexity, indicative of lower model performance. This is attributed to the fixed index potentially altering the original meaning of words. For instance, in the case of "New York City," unlearning the single token "City" would compel the model to disregard the frequent occurrence of "City" following "New York," consequently compromising the model's performance.

\textbf{Unlearning + Memory Implanting.}
To assess the effectiveness of sensitivity analysis coupled with selective forgetting in safeguarding the PII of targeted persons, we implemented unlearning in conjunction with memory implanting, as shown in Table~\ref{phone_ablation_table} ~\ref{address_ablation_table}. Our findings revealed that Unlearning + Memory Implanting proved capable of safeguarding the majority of persons' phone numbers and physical addresses, resulting in phone number risk scores of 0.2 and address risk scores of 6.7. However, this approach exhibited higher perplexity levels, measuring 16.3 and 33.6, which signifies diminished model performance. This is because the unlearning method essentially erases entire PII sequences, thereby enhancing PII protection capabilities at the expense of model performance.

\begin{wrapfigure}{r}{0.5\textwidth}
    \centering
    \includegraphics[width=0.5\textwidth, height=1.5\textwidth, keepaspectratio]{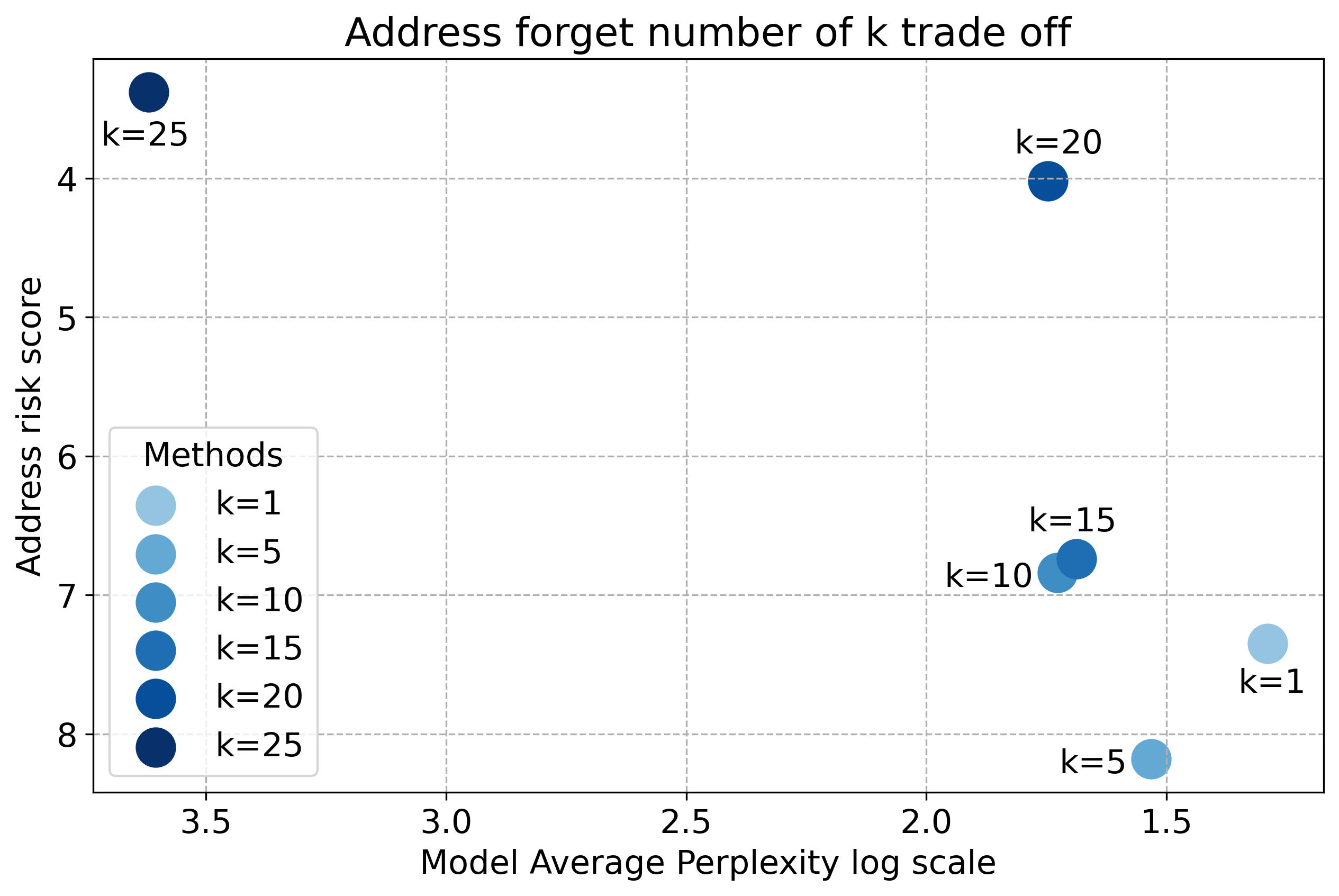}
  \caption{\rebuttalcaption{Address PPA Risk score vs forget number of indexes: PPA tunes the parameter $k$, as defined in Equations \ref{eq:topk} and \ref{eq:selected_unlearning}.}}
  \label{address_current_method_trade_off}
\vspace{-12pt}
\end{wrapfigure}

\subsection{PPA Trade-Offs: Number of forgotten indexes}
% Analysis of Address Mix Selective Unlearning forget number of indexes trade off
Table \ref{phone_enron_defense_table} and \ref{address_enron_defense_table} reveal that, after applying PPA, the address risk score remains at 7.3, while the phone risk score drops to 0. This disparity may be due to physical addresses being longer and less structured than phone numbers.  Therefore, we have conducted an ablation study to quantify the drop-off in risk score as the number of forgotten indexes increases. We conducted experiments on Address PPA. Specifically, we set \( k = 1, 5, 10, 15, 20, 25 \) in Equations \ref{eq:topk} and \ref{eq:selected_unlearning}. 
Both stages followed the training protocols in Appendix \ref{training_setting} for a single epoch.
For Addresses, the PPA method improves the PII risk score if more than one index is selected for forgetting. However, selecting too many indexes causes the model performance to deteriorate, as shown in Figure~\ref{address_current_method_trade_off}. This ablation demonstrates that PPA is a flexible method, allowing for adjustments to the balance between defense capability and model performance by modifying the number of key elements to be forgotten.

\section{Conclusion and Discussion\label{conclusion_and_discussion}}
We demonstrated that Proactive Privacy Amnesia achieves the optimal balance between defense performance and model utility compared to methods like Error Injection, Empty Response, Unlearning, and DEPN for protecting users' PII, including phone numbers and physical addresses. Additionally, we initially introduce the concept of the 'memorization factor', which affects the model's capacity to retain PII sequences. This concept is using in sensitivity analysis and supported by theoretical justification. Furthurmore, PPA is a flexible method that can adjust its balance between defense capability and model performance. Future work could extend PPA to protect the privacy of relationships, such as those between persons or between organizations. 

% Expand PPA to protect fields that contains privacy that people do not want to be exposed

\section{ACKNOWLEDGEMENTS\label{acknowledgement}}
This work is supported by ARO W911NF-23-2-0224, NSF IIS-2332744, and CNS-2112562. We also thank Accenture for its support and Yun-Chu Wang for her support. We thank area chair and reviewers for their valuable comments.

\bibliography{iclr2025_conference}
\bibliographystyle{iclr2025_conference}

\newpage
\appendix

{
\centering
\section*{Appendix}
}
\setcounter{proposition}{0}

\section{Existing Assets\label{existing_assets}}
The two existing assets used in this work are the Enron email dataset~\cite{klimt2004introducing} and aeslc dataset~\cite{zhang-tetreault-2019-email}. Please see their information below.

\begin{itemize}

\item Enron email dataset: The URL is http://www.enron-mail.com/email/; the
 license is not clearly stated by the authors.

\item aeslc dataset: The URL is https://huggingface.co/datasets/aeslc; the
 license is not clearly stated by the authors.

\end{itemize}

\section{Details of building ground truth table\label{build_ground_truth_table}}
To evaluate our defense method, we constructed a ground truth table comprising two parts: (1) We utilized the AWS Comprehend Service~\citep{aws2024comprehend} to extract PII, including names, phone numbers, and physical addresses; (2) we employed~\citep{manakul2023mqag} to determine the correlations between specific PIIs and the corresponding persons.

\section{Details of Evaluation Dataset\label{build_evaluation_dataset}}
For Enron email dataset, we constructed our evaluation ground truth table using the aelsc training dataset~\citep{zhang-tetreault-2019-email}. However, there is overlap between the aelsc training and validation datasets, which are used to train the soft prompt for the soft prompt attack. To ensure a fair comparison between soft prompt and probing attacks, we excluded certain persons' scores, as detailed in Appendix~\ref{details_enron_evaluation_dataset}. Consequently, our evaluation focused on 468 persons whose phone numbers were disclosed and 790 persons whose physical addresses were revealed.

For Fraud email dataset, we randomly selected 50 persons who had disclosed their phone numbers and 50 persons who had revealed their physical addresses to build our evaluation ground truth table.

\section{Evaluation Details of the Enron Email Experiment.\label{details_enron_evaluation_dataset}}
We constructed our evaluation ground truth table on the aeslc training dataset~\citep{zhang-tetreault-2019-email}, which comprises data from 1,359 persons. Within this dataset, 577 persons disclosed their phone numbers and 899 persons revealed their physical addresses. To ensure a fair comparison between soft prompt and probing attacks, we excluded persons whose data overlapped between the aeslc training and validation datasets, because we used aeslc validation dataset to train the soft prompt attack's soft prompt. The number of overlapping persons is 109. Consequently, our evaluation focused on 468 persons (577 - 109) whose phone numbers were exposed and 790 persons (899 - 109) whose physical addresses were exposed.
Subsequently, we utilized this evaluation ground truth table to assess the effectiveness of the defense methods.

\rebuttal{

\section{Comparison with Differential Privacy-Based methods \label{comparison_diff_privacy_method}}
We implemented the Differentially Private Decoding (DP Decoding) in ~\citep{majmudar2022differentially} and the Just Fine-Tune Twice (JFT) method in ~\citep{shi2022just}. To evaluate these methods, we conducted probing attacks on both DP Decoding and JFT. Specifically, for DP Decoding, the parameter $\lambda$ impacts both utility (the effectiveness of the generated output) and privacy (the protection of sensitive information). As $\lambda$ varies from 0 to 1, a value closer to 0 enhances privacy but reduces utility, while a value closer to 1 improves utility at the expense of privacy. We tested various values of the $\lambda$ parameter ranging from (0.1, 0.3, 0.5, 0.7, 0.9) and we selected $\lambda$=0.3 as it achieved the best balance between utility and privacy protection among the tested values, offering better utility compared to $\lambda$=0.1, and improved defense capability compared to $\lambda$=0.5, as shown in Table~\ref{tab:dp_decoding_lambda}. We observed that our PPA method still outperformed both DP Decoding and JFT, achieving a lower risk score and a higher utility score, as shown in Table~\ref{tab:comparison_diff_privacy_method}. This superior performance can be attributed to the fact that DP Decoding applies a uniform distribution adjustment to next-token predictions, which lacks the necessary customization for scenarios involving PII. 

\begin{table}[ht]
\renewcommand{\arraystretch}{1.5} % Adjust 
\resizebox{\textwidth}{!}{%
\centering
\begin{tabular}{lccccccc}
\hline
  & \textbf{PPA} & \textbf{$\lambda$=0.1} & \cellcolor[HTML]{EFEFEF} \textbf{$\lambda$=0.3} & \textbf{$\lambda$=0.5} & \textbf{$\lambda$=0.7} & \textbf{$\lambda$=0.9} & \textbf{no defense} \\ \hline
Risk Score \rebuttalcaption{\(\downarrow\)}          & 0.0 & 15.7 & \cellcolor[HTML]{EFEFEF} 30.4   & 32.4 & 33.2 & 41.4     & 57.9 \\ \hline
Exact Match Score \rebuttalcaption{\(\downarrow\)}   & 0.0 & 10.6 & \cellcolor[HTML]{EFEFEF} 28.4   & 30.1 & 31.1 & 38.6     & 56.0 \\ \hline
GPT-4o Email Score  & 5.2 & 3.0 (3.025) & \cellcolor[HTML]{EFEFEF} 4.8 (4.750) & 4.8 (4.825) & 4.9 (4.875) & 5.1 (5.125)  & 5.0 \\ \hline
\end{tabular}}
\caption{\rebuttalcaption{Comparison of Different $\lambda$ Values in DP Decoding. $\lambda$=0.3 achieved the best balance between utility and privacy protection among the tested $\lambda$ value.}}
\label{tab:dp_decoding_lambda}
\end{table}

\begin{table}[ht]
\renewcommand{\arraystretch}{1.5} % Adjust 
% \resizebox{\textwidth}{!}{%
\centering
\begin{tabular}{lccc}
\hline
\textbf{Phone Defense Model} & \textbf{Risk Score} \rebuttalcaption{\(\downarrow\)} & \textbf{Exact Match Score} \rebuttalcaption{\(\downarrow\)} & \textbf{GPT-4o Email Score} \\ \hline
DP Decoding & 30.4 & 28.4 & 4.8 \\ \hline
JFT & 28.4 & 26.0 & 5.0 \\ \hline
\cellcolor[HTML]{EFEFEF}PPA & \cellcolor[HTML]{EFEFEF}0.0 & \cellcolor[HTML]{EFEFEF}0.0 & \cellcolor[HTML]{EFEFEF}5.2 \\ \hline
\end{tabular}%}
\caption{\rebuttalcaption{Performance comparison of Differential Privacy-Based methods.}}
\label{tab:comparison_diff_privacy_method}
\end{table}

\section{Model performance evaluation on MMLU and TruthfulQA \label{truthfulqa_metric}}
We provide some new evaluations on the model's performance metrics on both MMLU~\citep{hendrycks2020measuring} and TruthfulQA~\citep{lin2021truthfulqa}. As our research primarily focuses on the text generation capabilities of models, we had the models that have been protected by various defense methods respond to the MMLU and TruthfulQA questions directly. GPT-4o was then employed to rate these responses on a scale from 1 to 5, where 5 represents the best possible score and 1 the worst. Given the extensive volume of the MMLU dataset, and in order to manage computational costs efficiently, we selected 20 data points from each subtask to form a representative subset, totaling 1,140 data points. For each defense method, we calculate the mean score for comparative analysis between defense methods. We found that PPA achieves the highest MMLU and TruthfulQA score among all baseline defense methods, as illustrated in Table~\ref{tab:truthfulqa_defense_comparison}.
\begin{table}[ht]
\renewcommand{\arraystretch}{1.5} % Adjust 
%\resizebox{\textwidth}{!}{%
\centering
\begin{tabular}{lccc}
\hline
\textbf{Phone Defense Model} & \textbf{MMLU Score} & \textbf{TruthfulQA Score} & \textbf{GPT-4o Email Score} \\ \hline
Empty Response & 3.3 & 3.4 & 5.7 \\ \hline
Error Injection & 3.3 & 3.2 & 5.2 \\ \hline
Unlearning & 1.6 & 1.7 & 1.1\\ \hline
DEPN & 2.3 & 2.4 & 2.0 \\ \hline
\cellcolor[HTML]{EFEFEF}PPA & \cellcolor[HTML]{EFEFEF}4.2 & \cellcolor[HTML]{EFEFEF}4.1 & \cellcolor[HTML]{EFEFEF} 5.2\\ \hline
\end{tabular}%}
\caption{\rebuttalcaption{Comparison of Phone Defense Models based on MMLU and TruthfulQA mean scores and GPT-4o Email Scores.}}
\label{tab:truthfulqa_defense_comparison}
\end{table}

\begin{comment}
\begin{table}[ht]
\renewcommand{\arraystretch}{1.5} % Adjust 
%\resizebox{\textwidth}{!}{%
\centering
\begin{tabular}{lcc}
\hline
\textbf{Phone Defense Model} & \textbf{TruthfulQA mean score} \\ \hline
\textbf{Empty Response} & 3.4 \\ \hline
\textbf{Error Injection} & 3.2 \\ \hline
\textbf{Unlearning} & 1.7\\ \hline
\textbf{DEPN} & 2.4 \\ \hline
\textbf{PPA} & 4.1\\ \hline
\end{tabular}%}
\caption{\rebuttalcaption{Comparison of Phone Defense Models based on TruthfulQA mean scores and GPT-4o Email Scores.}}
\label{tab:truthfulqa_defense_comparison}
\end{table}
\end{comment}

\begin{comment}
\begin{table}[ht]
\renewcommand{\arraystretch}{1.5} % Adjust 
\resizebox{\textwidth}{!}{%
\centering
\begin{tabular}{lcc}
\hline
\textbf{Phone Defense Model} & \textbf{TruthfulQA mean score} & \textbf{GPT-4o Email Score} \\ \hline
\textbf{Empty Response} & 3.4 & 5.7 \\ \hline
\textbf{Error Injection} & 3.2 & 5.2 \\ \hline
\textbf{Unlearning} & 1.7 & 1.1 \\ \hline
\textbf{DEPN} & 2.4 & 2.0 \\ \hline
\textbf{PPA} & 4.1 & 5.2 \\ \hline
\end{tabular}}
\caption{Comparison of Phone Defense Models based on TruthfulQA mean scores and GPT-4o Email Scores.}
\label{tab:truthfulqa_defense_comparison}
\end{table}
\end{comment}

\section{Stronger Attacker has prior knowledge of the PII \label{stronger_prior_knowledge}}
We have implemented a more advanced attack scenario where the attacker possesses prior knowledge of the PII. Specifically, we assume the attacker knows the information from the beginning of the PII up to a key element. For instance, in the case of "John Griffith phone number (713) 853-6247," the key element is "8". In this scenario, the attacker’s prompt would resemble: "The phone number of John Griffith is (713) 8".

As shown in the Table~\ref{tab:strong_attacker_prior_knowledge}, we observe that PPA achieves the best balance between defense capability and model performance.

\begin{table}[ht]
\renewcommand{\arraystretch}{1.5} % Adjust 
% \resizebox{\textwidth}{!}{%
\centering
\begin{tabular}{lccc}
\hline
\textbf{Phone Defense Model} & \textbf{Risk Score} \rebuttalcaption{\(\downarrow\)} & \textbf{Exact Match Score} \rebuttalcaption{\(\downarrow\)} & \textbf{GPT-4o Email Score} \\ \hline
Empty Response & 154.0 & 141.5 & 5.7 \\ \hline
Error Injection & 75.1 & 69.1 & 5.2 \\ \hline
Unlearning & 6.7 & 1.8 & 1.1 \\ \hline
DEPN & 36.2 & 27.3 & 2.0 \\ \hline
\cellcolor[HTML]{EFEFEF}PPA & \cellcolor[HTML]{EFEFEF}12.1 & \cellcolor[HTML]{EFEFEF}9.8 & \cellcolor[HTML]{EFEFEF}5.2 \\ \hline
\end{tabular}%}
\caption{\rebuttalcaption{Phone Defense Models against strong attacker has the prior knowledge.}}
\label{tab:strong_attacker_prior_knowledge}
\end{table}

\section{Details of the proportions of key elements\label{key_elements_proportions}}
We calculated the proportions of key element lengths relative to the total lengths for phone numbers and physical addresses, which are 6.7\% and 27.6\%, respectively.

\section{More discussion about Memory Implanting\label{discuss_memory_implanting}}
We modified the memory implanting component to focus on replacing the key element with a different token. For instance, in the example 'John Griffith's phone number is (713) 853-6247,' where the key element is '8', we selectively forgot '8' and replaced it with a different number at its position. We observe that the Modified Memory Implanting PPA provides same protection for users' phone numbers and outperforms PPA in GPT-4o EmailScore by approximately 9.6\%, as shown in Table~\ref{tab:modified_memory_implanting}. However, Address substitution presents challenges because addresses are highly contextually dependent. Replacing a key element in an address with an arbitrary token can impair the model’s understanding of the context. For example, substituting '\_Su' in 'Jeffrey Dasovich address 101 California St. Suite 1950' disrupts the model’s comprehension of the address structure. Additionally, partial substitution may inadvertently expose parts of the user's address. Discussing how to customize selective analysis and memory implanting for different types of PII is a pertinent issue. Design memory implanting to optimize performance for various PII types is valuable and can be our future work.

\begin{table}[ht]
\renewcommand{\arraystretch}{1.5} % Adjust 
%\resizebox{\textwidth}{!}{%
\centering
\begin{tabular}{lccc}
\hline
\textbf{Phone Defense Model} & \textbf{Risk Score} \rebuttalcaption{\(\downarrow\)} & \textbf{Exact Match Score} \rebuttalcaption{\(\downarrow\)} & \textbf{GPT-4o Email Score} \\ \hline
Empty Response & 37.2 & 34.8 & 5.7 \\ \hline
Error Injection & 19.3 & 17.6 & 5.2 \\ \hline
Unlearning & 0.0 & 0.0 & 1.1 \\ \hline
DEPN & 0.0 & 0.0 & 2.0 \\ \hline
PPA & 0.0 & 0.0 & 5.2 \\ \hline
\cellcolor[HTML]{EFEFEF}Modified Memory Implanting PPA & \cellcolor[HTML]{EFEFEF}0.0 & \cellcolor[HTML]{EFEFEF}0.0 & \cellcolor[HTML]{EFEFEF}5.7 \\ \hline
\end{tabular}%}
\caption{\rebuttalcaption{Comparison of the Modified Memory Implanting PPA with Other Phone Defense Strategies.}}
\label{tab:modified_memory_implanting}
\end{table}

\section{Adding the exposure metric\label{exposure_metric}}

We calculated the exposure metric~\citep{carlini2019secret} for all baseline methods. Since calculating the exposure of PII is computationally intensive, we followed the approach in Table 2 of ~\citep{carlini2019secret} and evaluated the exposure for 10 phone numbers. Our results show that PPA outperforms other baseline defense methods, as shown in Table~\ref{tab:exposure_metric}
\begin{table}[ht]

%\begin{wraptable}{r}{0.45\textwidth} % 'r' for right side and width of the wraptable
\renewcommand{\arraystretch}{1.5} % Adjust 
%\resizebox{\textwidth}{!}{%
\centering
\begin{tabular}{lc}
\hline
\textbf{Phone Defense Model} & \textbf{Exposure} \\ \hline
Empty Response & 12.50 \\ \hline
Error Injection & 10.94 \\ \hline
Unlearning & 3.55 \\ \hline
DEPN & 7.72 \\ \hline
\cellcolor[HTML]{EFEFEF}PPA & \cellcolor[HTML]{EFEFEF}0.05 \\ \hline
\end{tabular}%}
\caption{\rebuttalcaption{Exposure levels of various Phone Defense Strategies.}}
\label{tab:exposure_metric}
%\end{wraptable}
\end{table}

\section{Evaluation on email address.\label{evaluate_email_address}}
We conducted an additional experiment to evaluate the protection of 281 users' email addresses in the aeslc training dataset. Using Levenshtein distance~\citep{po2020similarity}, we compared the predicted email addresses to the ground truth. As shown in the Table~\ref{tab:email_defense_model_scores}, PPA successfully defends all users' email addresses against probing attacks while maintaining model performance comparable to other baseline defense methods.
\begin{table}[ht]
\renewcommand{\arraystretch}{1.5} % Adjust 
%\resizebox{\textwidth}{!}{%
\centering
\begin{tabular}{lccc}
\hline
\textbf{Email Defense Model} & \textbf{Risk Score} \rebuttalcaption{\(\downarrow\)} & \textbf{Exact Match Score} \rebuttalcaption{\(\downarrow\)} & \textbf{GPT-4o Email Score} \\ \hline
Empty Response & 47.2 & 40.5 & 5.1 \\ \hline
Error Injection & 19.6 & 17.0 & 5.3 \\ \hline
Unlearning & 1.0 & 1.0 & 1.6 \\ \hline
DEPN & 1.0 & 1.0 & 1.3 \\ \hline
\cellcolor[HTML]{EFEFEF}PPA & \cellcolor[HTML]{EFEFEF}0.0 & \cellcolor[HTML]{EFEFEF}0.0 & \cellcolor[HTML]{EFEFEF}5.0 \\ \hline
\end{tabular}%}
\caption{\rebuttalcaption{Comparative Analysis of Email Defense Strategies Against Various Attacks in Enron Email Experiment.}}
\label{tab:email_defense_model_scores}
\end{table}

\section{Originally safe information to be exposed?\label{original_safe_expose}}
Motivated by the concern that the PPA defense could inadvertently expose previously secure PII of users who are not explicitly protected by the method.
We evaluated the exposure metric~\citep{carlini2019secret} for safe phone numbers—those not exposed to attackers—that were not protected by the PPA method, using both the no-defense setup and the PPA model. Given the time-intensive nature of calculating PII exposure, we referenced Table 2 from The Secret Sharer ~\citep{carlini2019secret} and analyzed the exposure of 10 such phone numbers. The average exposure for these cases is summarized in the Table~\ref{tab:exposure_other_user}.

As shown in the Table~\ref{tab:exposure_other_user}, the exposure of phone numbers not protected by the PPA method decreases slightly, from 1.57 (no defense) to 1.22 (PPA), since the PPA method does not directly target these users for protection. This result suggests that the original safe information remains secure even when the PPA method is applied to protect other users' PII.

\begin{table}[ht]

%\begin{wraptable}{r}{0.45\textwidth} % 'r' for right side and width of the wraptable
\renewcommand{\arraystretch}{1.5} % Adjust 
%\resizebox{\textwidth}{!}{%
\centering
\begin{tabular}{lc}
\hline
\textbf{Phone Defense Model} & \textbf{Exposure} \\ \hline
No Defense & 1.57 \\ \hline
PPA & 1.22 \\ \hline
\end{tabular}%}
\caption{\rebuttalcaption{Exposure of Users' Phone Numbers not protected by the PPA Method.}}
\label{tab:exposure_other_user}
%\end{wraptable}
\end{table}

\section{Discuss scalability for PPA\label{PPA_scalability}}

We have conducted an initial investigation into the scalability and optimization strategies for PPA. Our experiments involved combining PPA with efficient fine-tuning techniques, such as LoRA ~\citep{hu2021lora}, using a rank of 16 and an alpha value of 32. As shown in the Table~\ref{tab:ppa_lora}, applying LoRA to PPA produced promising results: after fine-tuning for three epochs, the risk score reduced to 1.0, and after four epochs, it further decreased to 0.0, all while maintaining comparable model performance. Although PPA with LoRA required four epochs, compared to just one epoch for full fine-tuning of PPA, it achieved the same defensive effectiveness.

Table~\ref{tab:ppa_lora} demonstrates that PPA has potential for scalability. Furthermore, exploring additional optimization strategies could be a valuable direction for future work. 

\begin{table}[ht]

%\begin{wraptable}{r}{0.45\textwidth} % 'r' for right side and width of the wraptable
\renewcommand{\arraystretch}{1.5} % Adjust 
%\resizebox{\textwidth}{!}{%
\centering
\begin{tabular}{lccc}
\hline
\textbf{Phone Defense Model} & \textbf{Risk Score \rebuttalcaption{\(\downarrow\)}} & \textbf{Exact Match Score \rebuttalcaption{\(\downarrow\)}} & \textbf{GPT-4o Email Score} \\ \hline
PPA LoRA 1-epoch & 24.6 & 23.6 & 5.4 \\ \hline
PPA LoRA 2-epoch & 5.4 & 5.2 & 5.2 \\ \hline
PPA LoRA 3-epoch & 1.0 & 1.0 & 5.0 \\ \hline
PPA LoRA 4-epoch & 0.0 & 0.0 & 5.1 \\ \hline
\end{tabular}%}
\caption{\rebuttalcaption{Comparative Analysis of PPA LoRA with different fine-tuning epochs.}}
\label{tab:ppa_lora}
%\end{wraptable}
\end{table}

\section{Ablation study on various forgetting strategies\label{forgetting_strategies}}
We conducted ablation studies on protecting the user's physical address using the following three strategies: 1) Only the most sensitive tokens (PPA), 2) Span around the most sensitive token, and 3) The most sensitive tokens + the following tokens. Their respective Risk Scores and GPT-4o Email Scores are presented in Table~\ref{tab:forget_strategies}.

As shown in the Table~\ref{tab:forget_strategies}, "Only the most sensitive token (PPA)" achieves an optimal balance between defense capability and model performance. "Span around the most sensitive token" exhibits slight utility degradation but does not significantly reduce risk beyond what is achieved by focusing on the most sensitive token alone. "The most sensitive tokens + the following tokens" strategy reduces the risk score further but at the cost of significantly lowering the GPT-4o Email Score. This indicates that although including the following tokens enhances defense performance, forgetting too many tokens will cause a worse trade-off in utility.

Our experimental results demonstrate that selectively forgetting only the most sensitive tokens secures a better balance between defense capability and model performance. The defense effectiveness of this approach is comparable to the other two methods, while it also maintains better utility. Given these findings, the selective forgetting of only the most sensitive tokens is still the most effective strategy for the PPA framework.

\begin{table}[ht]

%\begin{wraptable}{r}{0.45\textwidth} % 'r' for right side and width of the wraptable
\renewcommand{\arraystretch}{1.5} % Adjust 
%\resizebox{\textwidth}{!}{%
\centering
\begin{tabular}{lcc}
\hline
\textbf{Forgetting Strategies} & \textbf{Risk Score \rebuttalcaption{\(\downarrow\)}} & \textbf{GPT-4o Email Score} \\ \hline
Only the most sensitive tokens (PPA)              & 4.7 (4.702) &  3.6 \\ \hline
Span around the most sensitive token              & 4.7 (4.737)  &  3.4 \\ \hline
The most sensitive tokens + the following tokens  & 4.4          &  2.9 \\ \hline
\end{tabular}%}
\caption{\rebuttalcaption{Comparative Analysis of various forgetting strategies.}}
\label{tab:forget_strategies}
%\end{wraptable}
\end{table}
}

\section{Details of model performance perplexity metric\label{perplexity_details}}

In Tables~\ref{phone_enron_defense_table}, ~\ref{address_enron_defense_table}, and \ref{fraud_defense_table}, we report the average Perplexity across three different tests to evaluate model performance. The detailed results of each individual Perplexity test are provided in Tables\ref{phone_perplexity_enron_table}, ~\ref{address_perplexity_enron_table}, and ~\ref{perplexity_fraud_table}.

% \usepackage{multirow}
\begin{table}[t]
\renewcommand{\arraystretch}{1.55} % Adjust 
\resizebox{\textwidth}{!}{%
\centering
\begin{tabular}{ll|cccc}
\hline
\multicolumn{2}{l|}{}                                                                                                                & \multicolumn{4}{c}{\textbf{Model Performance}}                                                                                                                                        \\ \cline{3-6} 
\multicolumn{2}{l|}{\multirow{-2}{*}{\textbf{\begin{tabular}[c]{@{}l@{}}Enron Email Experiment\\ Phone Number Defense Model\end{tabular}}}} & Enron perplexity first=512 tokens     & Enron perplexity stride=256           & GPT4 perplexity                        & \begin{tabular}[c]{@{}c@{}}Perplexity\\ \end{tabular} \\ \hline
\multicolumn{1}{c|}{}                                                  & Original                                                    & 7.6                                   & 3.2                                   & 6.2                                    & 5.6                                                          \\ 
\multicolumn{1}{c|}{}                                                  & Finetuned                                                   & 26.1                                  & 6.6                                   & 16.0                                   & 16.2                                                         \\ \cline{2-6} 
\multicolumn{1}{c|}{}                                                  & Empty Response                                              & 26.7                                  & 6.7                                   & 16.2                                   & 16.5                                                         \\ 
\multicolumn{1}{c|}{}                                                  & Error Injection                                             & 23.2                                  & 6.3                                   & 14.2                                   & 14.6                                                         \\ 
\multicolumn{1}{c|}{}                                                  & Unlearning                                                  & $9.7\times10^{11}$                    & 5691                                  & $1.9\times10^{6}$                      & $3.2\times10^{11}$                                           \\ 
\multicolumn{1}{c|}{}                                                  & DEPN                                                        & 139.5                                 & 19.9                                  & 72.3                                   & 77.2                                                         \\ 
\multicolumn{1}{c|}{\multirow{-7}{*}{LLaMA2-7b}}                       & \cellcolor[HTML]{EFEFEF}PPA                                & \cellcolor[HTML]{EFEFEF}26.7          & \cellcolor[HTML]{EFEFEF}6.8           & \cellcolor[HTML]{EFEFEF}14.4           & \cellcolor[HTML]{EFEFEF}16.0                                               \\ \hline
\multicolumn{1}{l|}{}                                                  & Original                                                    & 10.5                                  & 4.4                                   & 15.0                                   & 9.9                                                          \\ 
\multicolumn{1}{l|}{}                                                  & Finetuned                                                   & 58.9                                  & 11.4                                  & 167.4                                  & 79.2                                                         \\ \cline{2-6} 
\multicolumn{1}{l|}{}                                                  & Empty Response                                              & 61.7                                  & 11.7                                  & 175.5                                  & 82.9                                                         \\ 
\multicolumn{1}{l|}{}                                                  & Error Injection                                             & 46.5                                  & 9.9                                   & 125.2                                  & 60.5                                                         \\ 
\multicolumn{1}{l|}{}                                                  & Unlearning                                                  & $1.5\times10^{22}$                    & $8.5\times10^{14}$                    & $1.1\times10^{14}$                     & $5.0\times10^{21}$                                           \\ 
\multicolumn{1}{l|}{}                                                  & DEPN                                                        & 92.2                                  & 14.9                                  & 308.6                                  & 138.5                                                        \\ 
\multicolumn{1}{l|}{\multirow{-7}{*}{LLaMA3-8b}}                       & \cellcolor[HTML]{EFEFEF}PPA                                & \cellcolor[HTML]{EFEFEF}57.5          & \cellcolor[HTML]{EFEFEF}11.3          & \cellcolor[HTML]{EFEFEF}133.8          & \cellcolor[HTML]{EFEFEF}67.5                                              
\end{tabular}%
}
\caption{\label{phone_perplexity_enron_table}Comparative Analysis of Perplexity Against Various Attacks in Enron Email Experiment. PPA stands for Proactive Privacy Amnesia. 'Perplexity' refers to the average of our perplexity metric.}
\vspace{-12pt}
\end{table}

% \usepackage{multirow}
\begin{table}[t]
\renewcommand{\arraystretch}{1.55} % Adjust 
\resizebox{\textwidth}{!}{%
\centering
\begin{tabular}{ll|cccc}
\hline
\multicolumn{2}{l|}{}                                                                                                                  & \multicolumn{4}{c}{\textbf{Model Performance}}                                                                                                                                       \\ \cline{3-6} 
\multicolumn{2}{l|}{\multirow{-2}{*}{\textbf{\begin{tabular}[c]{@{}l@{}}Enron Email Experiment\\ Physical Address Defense Model\end{tabular}}}} & Enron perplexity first=512 tokens     & Enron perplexity stride=256           & GPT4 perplexity                       & \begin{tabular}[c]{@{}c@{}}Perplexity\end{tabular} \\ \hline
\multicolumn{1}{c|}{}                                                   & Original                                                     & 7.6                                   & 3.2                                   & 6.2                                   & 5.6                                                          \\ 
\multicolumn{1}{c|}{}                                                   & Finetuned                                                    & 26.1                                  & 6.6                                   & 16.0                                  & 16.2                                                         \\ \cline{2-6} 
\multicolumn{1}{c|}{}                                                   & Empty Response                                               & 27.3                                  & 6.7                                   & 16.1                                  & 16.7                                                         \\ 
\multicolumn{1}{c|}{}                                                   & Error Injection                                              & 23.5                                  & 6.4                                   & 13.3                                  & 14.4                                                         \\ 
\multicolumn{1}{c|}{}                                                   & Unlearning                                                   & inf                                   & $3.1\times10^{31}$                    & $4.73\times10^{21}$                   & inf                                                          \\ 
\multicolumn{1}{c|}{}                                                   & DEPN                                                         & 480.4                                 & 43.8                                  & 132.6                                 & 218.9                                                        \\ 
\multicolumn{1}{c|}{\multirow{-7}{*}{LLaMA2-7b}}                        & \cellcolor[HTML]{EFEFEF}PPA                        & \cellcolor[HTML]{EFEFEF}35.6 & \cellcolor[HTML]{EFEFEF}8.2  & \cellcolor[HTML]{EFEFEF}14.6 & \cellcolor[HTML]{EFEFEF}19.5                        \\ \hline
\multicolumn{1}{l|}{}                                                   & Original                                                     & 10.5                                  & 4.4                                   & 15.0                                  & 9.9                                                          \\ 
\multicolumn{1}{l|}{}                                                   & Finetuned                                                    & 58.9                                  & 11.4                                  & 167.4                                 & 79.2                                                         \\ \cline{2-6} 
\multicolumn{1}{l|}{}                                                   & Empty Response                                               & 60.0                                  & 11.5                                  & 165.1                                 & 78.8                                                         \\ 
\multicolumn{1}{l|}{}                                                   & Error Injection                                              & 44.8                                  & 9.7                                   & 103.2                                 & 52.5                                                         \\ 
\multicolumn{1}{l|}{}                                                   & Unlearning                                                   & $1.2\times10^{27}$                    & $1.3\times10^{18}$                    & inf                                   & inf                                                          \\ 
\multicolumn{1}{l|}{}                                                   & DEPN                                                         & 177.8                                 & 23.2                                  & 403.7                                 & 201.5                                                        \\ 
\multicolumn{1}{l|}{\multirow{-7}{*}{LLaMA3-8b}}                        & \cellcolor[HTML]{EFEFEF}PPA                        & \cellcolor[HTML]{EFEFEF}63.0 & \cellcolor[HTML]{EFEFEF}12.4 & \cellcolor[HTML]{EFEFEF}97.4 & \cellcolor[HTML]{EFEFEF}57.6                       
\end{tabular}%
}
\caption{\label{address_perplexity_enron_table}Comparative Analysis of Perplexity Against Various Attacks in Enron Email Experiment. PPA stands for Proactive Privacy Amnesia. 'Perplexity' refers to the average of our perplexity metric.}
\vspace{-12pt}
\end{table}

\begin{table}[t]
\renewcommand{\arraystretch}{1.5} % Adjust 
\resizebox{\textwidth}{!}{%
\centering
\begin{tabular}{cl|ccc}
\hline
\multicolumn{2}{l|}{}                                                                                                                              & \multicolumn{3}{c}{\textbf{Model Performance}}                                                                                               \\ \cline{3-5} 
\multicolumn{2}{l|}{\multirow{-2}{*}{\textbf{\begin{tabular}[c]{@{}l@{}}Fraud Email Experiment\\ Defense Model\end{tabular}}}}                     & Enron perplexity first=512 tokens     & Enron perplexity stride=256           & \begin{tabular}[c]{@{}c@{}}Perplexity\end{tabular} \\ \hline
\multicolumn{1}{c|}{}                                                                                      & Original                              & 5.33                                  & 2.79                                  & 4.06                                                         \\ 
\multicolumn{1}{c|}{}                                                                                      & Finetuned                             & 1.17                                  & 1.05                                  & 1.11                                                         \\ \cline{2-5} 
\multicolumn{1}{c|}{}                                                                                      & Empty Response                        & 1.17                                  & 1.05                                  & 1.11                                                         \\ 
\multicolumn{1}{c|}{}                                                                                      & Error Injection                       & 1.15                                  & 1.05                                  & 1.10                                                         \\ 
\multicolumn{1}{c|}{}                                                                                      & Unlearning                            & 1.53                                  & 1.14                                  & 1.33                                                         \\ 
\multicolumn{1}{c|}{}                                                                                      & DEPN                                  & 1.27                                  & 1.09                                  & 1.18                                                         \\ 
\multicolumn{1}{c|}{\multirow{-7}{*}{\begin{tabular}[c]{@{}c@{}}LLaMA2-7b\\ Phone Number Defense\end{tabular}}}   & \cellcolor[HTML]{EFEFEF}PPA & \cellcolor[HTML]{EFEFEF}1.16 & \cellcolor[HTML]{EFEFEF}1.05 & \cellcolor[HTML]{EFEFEF}1.10                        \\ \hline
\multicolumn{1}{c|}{}                                                                                      & Original                              & 5.33                                  & 2.79                                  & 4.06                                                         \\ 
\multicolumn{1}{c|}{}                                                                                      & Finetuned                             & 1.17                                  & 1.05                                  & 1.11                                                         \\ \cline{2-5} 
\multicolumn{1}{c|}{}                                                                                      & Empty Response                        & 1.17                                  & 1.05                                  & 1.11                                                         \\ 
\multicolumn{1}{c|}{}                                                                                      & Error Injection                       & 1.15                                  & 1.05                                  & 1.10                                                         \\ 
\multicolumn{1}{c|}{}                                                                                      & Unlearning                            & 48.21                                 & 1.14                                  & 1.33                                                         \\ 
\multicolumn{1}{c|}{}                                                                                      & DEPN                                  & 2.48                                  & 1.09                                  & 1.18                                                         \\ 
\multicolumn{1}{c|}{\multirow{-7}{*}{\begin{tabular}[c]{@{}c@{}}LLaMA2-7b\\ Physical Address Defense\end{tabular}}} & \cellcolor[HTML]{EFEFEF}PPA & \cellcolor[HTML]{EFEFEF}1.16 & \cellcolor[HTML]{EFEFEF}1.05 & \cellcolor[HTML]{EFEFEF}1.10                       
\end{tabular}%
}
\caption{\label{perplexity_fraud_table}Comparative Analysis of Perplexity Against Various Attacks in Enron Email Experiment. PPA stands for Proactive Privacy Amnesia. 'Perplexity' refers to the average of our perplexity metric.}
\vspace{-12pt}
\end{table}

\section{Details of the Baseline defense methods\label{baseline_defense_details}}
\textbf{Empty Response}~\citep{patil2023can, ouyang2022training}\textbf{.} This method refines the model to label non-sensitive information as "dummy". For instance, we create templates for each person, formatted with the person’s name, PII type, and "dummy". We then perform gradient descent on "dummy" following the training settings outlined in Appendix~\ref{training_setting} with a single epoch.

\textbf{Error Injection.} We implemented the Error Injection method on each person's phone numbers, conducting a single epoch of training. This same process is used to preserve a person's physical addresses. Take a person's phone number as an example, we create templates for each person, structured as the person’s name, PII type, and fake PII, which is generated by~\citep{presidioResearch2024}. We then apply gradient descent to false PII, adhering to the training settings detailed in Appendix~\ref{training_setting}.

\textbf{Unlearning}~\citep{jang2022knowledge}\textbf{.} We applied an unlearning technique to the PII sequence by performing gradient ascent on it, following the training settings specified in Appendix~\ref{training_setting} with a single epoch.

\textbf{DEPN}~\citep{wu2023depn}
We adopted the DEPN approach, as detailed in the DEPN GitHub repository, to protect PII, specifically phone numbers and physical addresses. Our goal was to eliminate specific neurons from the output of the LlamaDecoderLayer in the LlamaModel~\citep{meta_llama_2_7b}. We established a threshold ratio of 0.01 for both phone numbers and physical addresses, with mode ratio bags set at 0.49 and 0.5, respectively. Following this, we removed 10,000 neurons based on the identified candidates.

\section{Details of Attack methods\label{attack_details}}
For the input rephrasing attack, we generated 20 attack templates based on the twin template described in~\citep{kim2024propile}.

For soft prompt tuning in the Enron email experiment, we used the first probing twin template from~\citep{kim2024propile}, leveraging the aeslc validation ground truth table. In the Fraud email experiment, we applied the same probing template, selecting 25 persons with phone numbers and 25 with physical addresses randomly from the fraud email dataset.

\section{Details of Attack Success Metric\label{attack_success_metric_details}}
For the total phone numbers and physical address risk score, we calculate the average phone number risk score and average physical address risk score for each person. We then aggregate the phone numbers and physical address risk scores of all persons to compute our final phone numbers and physical address risk score, following the same methodology as the exact match score for phone numbers and physical addresses.

\section{Model Performance Metric\label{model_performance_metric_details}}
For the \textbf{Perplexity metric}, we conducted three different tests to assess model performance. First, we calculated perplexity~\citep{huggingface_perplexity} on the first 512 tokens of each text (with a maximum length of 512 tokens). Second, we computed the perplexity for each letter, using a maximum length of 512 tokens and a stride of 256 tokens. Third, we assessed the perplexity of letters generated by GPT-4~\citep{achiam2023gpt}, but the Fraud email experiment did not have this metric, because GPT-4 cannot write fraud emails due to its safety aligned mechanism. These three tests help us determine whether our defense method impacts model performance. 

For the \textbf{Email completion metric} in the Enron email experiment, we evaluated the model's performance in completing truncated emails. Specifically, we tasked the model with completing 40 truncated emails, which were subsequently evaluated by GPT-4o~\citep{openai2024gpt4o}. Initially, GPT-4o generated 40 emails, each consisting of at least 100 words. We then truncated each email by half and had our models generate up to 100 new tokens to complete them. GPT-4o assessed and ranked the completions on a scale of 1 to 10, with 10 representing the best score. The average score was calculated across all 40 completions.

For the \textbf{Email completion metric} in the Fraud email experiment, we evaluated the model's ability to generate complete fraud emails. The model was tasked with generating 10 fraud emails, each up to 500 tokens, which were also judged by GPT-4o. GPT-4o ranked these completions on the same 1 to 10 scale, with the average score being calculated across all 10 fraud email completions.

\section{Proof}
\subsection{Proof of Proposition \ref{proposition1}}
\begin{proposition} 
Maximizing the memorization factor can lead to
\begin{align}
    \max_k D(k) = \left\{
    \begin{array}{lll}
        \max_k L(k)&\text{if } \exists k, \nabla L(k)=0,    \\
        \max_k 1/d_{\text{Newton}}(k)& \text{if } \nexists k, \nabla L(k)=0.
    \end{array}
\right.
\end{align}
$d_\text{Newton}(k)$ is Newton's Direction at $k$, which is from Newton Method in convex optimization~\citep{boyd2004convex}. $\max_k 1/d_{\text{Newton}}(k)$ is achieved when $d_{\text{Newton}}(k)\rightarrow 0^+$. As $L(k)$ is monotonically non-decreasing, a small positive $d_\text{Newton}(k)$ implies that the gradient at token $k$ quickly decreases with a negative second-order derivative.
\begin{proof}
Notice that 
\begin{align}
    L(k) = &-\sum_{\vx_1\cdots,\vx_k} p(\vx_1\cdots,\vx_k)\log q(\vx_1\cdots,\vx_k)\label{eq:accumulate_H}\\
    =&-\sum_{\vx_1\cdots,\vx_{k-1}} p(\vx_1\cdots,\vx_{k-1})\log q(\vx_1\cdots,\vx_{k-1})\nonumber\\
    &-\sum_{\vx_1\cdots,\vx_{k-1}} p(\vx_1\cdots,\vx_{k-1})\sum_{\vx_k}p(\vx_k|\vx_1\cdots,\vx_{k-1})\log q(\vx_k|\vx_1\cdots,\vx_{k-1})\\
    =&L(k-1)+H_k,
\end{align}
So we have 
\begin{align}
    &H_k=L(k)-L(k-1)\approx \nabla L(k),\\
    &H_{k+1}-H(k) \approx \nabla L(k+1)-\nabla L(k)\approx \nabla^2 L(k),\\
    &D_k = \frac{H_k-H_{k+1}}{H_k} \approx -\frac{\nabla^2 L(k)}{\nabla L(k)}.
\end{align}
Our selection method selects $k$ with the largest $D_k$. We discuss it in two situations:
\begin{enumerate}
    \item When there exists $k$ such that $H_k=\nabla L(k)=0$, we require that $\nabla^2L(k)<0$ to achieve the maximum ($D_k=+\infty$), this guarantees that $k$ achieves the maximum of $L(k)$ as well.
    \item When $H_k$ is always positive (notice that $H_k$ is never negative), $L(k)$ keeps growing as $k$ increases so we cannot find the maximum. But we still have
    \begin{align}
        \max_k D_k = \max_k \frac{1}{d_{\text{Newton}}(k)}=\min_k d_{\text{Newton}}(k),
        \label{eq:newton_direction}
    \end{align}
    where $d_{\text{Newton}}(k)=-\nabla L(k)/\nabla^2L(k)$ is \textit{Newton's Direction} in the second-order Newton's Method. The maximization is achieved when $d_{\text{Newton}}(k)\rightarrow 0^+$. Since $\nabla L(k)> 0$, $d_{\text{Newton}}(k)\rightarrow 0^+$ is achieved when $\nabla^2 L(k)=-\infty$, which implies that the gradient at $k$ quickly approaches $0$. 
\end{enumerate}
\end{proof}
\end{proposition}

\section{Training setting and hardware\label{training_setting}}
The training settings for fine-tuning the LLM on the Enron and Fraud email datasets, as well as for implementing defensive methods such as gradient descent and ascent, are as follows: a batch size of 4, the AdamW optimizer, a learning rate of 5e-5, weight decay of 0.001, a cosine learning rate scheduler, and a warmup ratio of 0.03. All experiments were conducted using 8 NVIDIA Quadro RTX 6000 24GB GPUs.

\begin{comment}
\section{Broader Impacts\label{boarder_impacts}}

Our work proposes a new method for preserving PII in LLMs, and we will explore the societal implications of techniques used for safeguarding PII.

\textbf{Positive impacts.} Safeguarding PII in LLMs while preserving their performance has significant positive implications. For instance, it can protect PII from fine-tuned LLMs with only a minimal drop in performance, ensuring that the LLMs remain effective for their intended purposes.

\textbf{Negative impacts.} Dynamic Mix Selected Unlearning consists of three stages: sensitivity analysis, selected unlearning, and error injection. These stages are unlikely to cause significant negative societal impacts.
\end{comment}

% \section{\JY{Additional Results? Please find a name for this section}}

\begin{figure*}[t!]
\centering
  \includegraphics[width=0.81\textwidth]{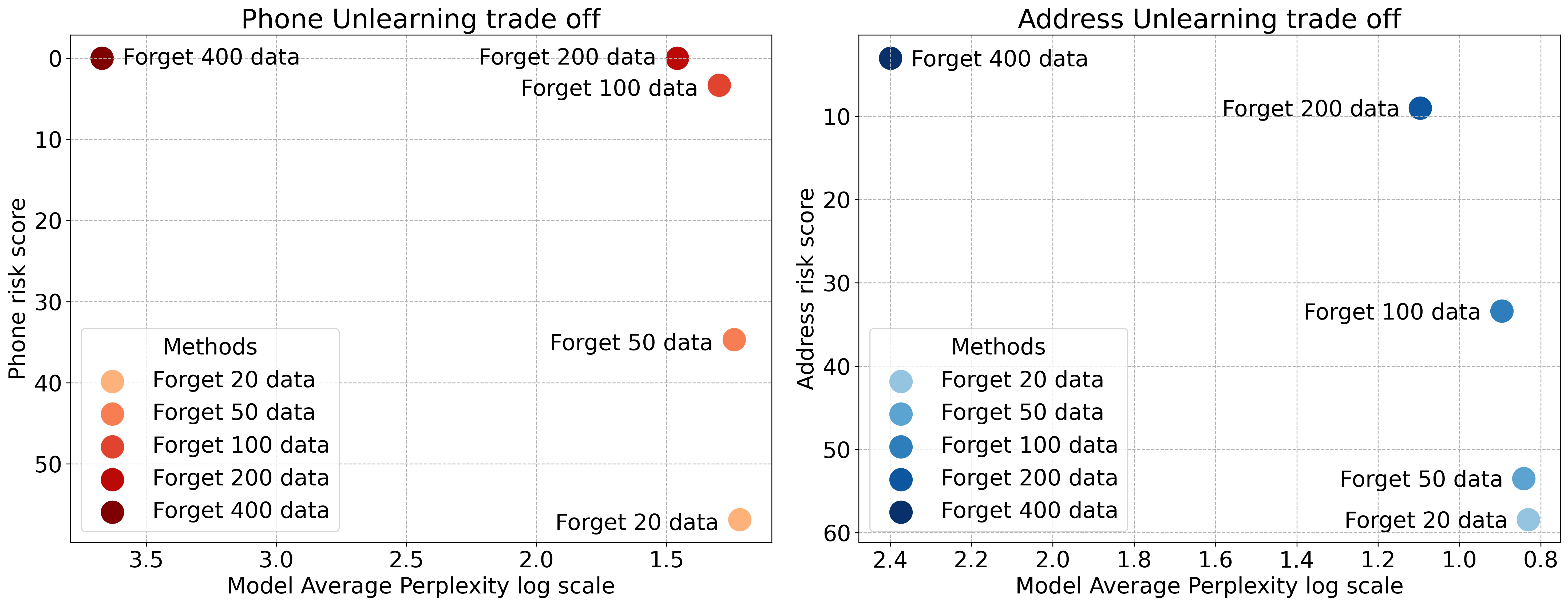}
  \caption{ Unlearning method trade-off: Risk score vs forget number of data. left: phone numbers; right: physical addresses }
  \label{phone_address_unlearning_break_even_trade_off}
\end{figure*}

\section{Additional analysis on unlearning scaling experiment}

\subsection{Unlearning method trade-off}
To analyze the break-even point of the unlearning method, we conducted experiments focusing on both phone numbers and address unlearning. We tested the forgetting of 20, 50, 100, 200, and 400 data points. The results indicate that as more data points are forgotten, a greater number of phone numbers and physical addresses are preserved. However, this leads to a deterioration in the model's performance, as illustrated in Figure~\ref{phone_address_unlearning_break_even_trade_off}. We discovered that forgetting between 200 and 400 data points significantly increases perplexity and it indicated that the break-even point for the unlearning method is when between 200 and 400 data points are forgotten.

\section{Broader Impacts.}
The societal implications of our work include positive impacts, as it can protect PII from fine-tuned LLMs with only a negligible drop in performance, ensuring that the LLMs remain effective for their intended purposes. And it is unlikely to cause significant negative societal impacts.

\iffalse
\appendix
\section{Appendix}
You may include other additional sections here.
\fi

\end{document}